\documentclass[11pt]{article}

\newif\ifanonymouspaper
\ifdefined\ANONYMOUSVERSION
  \anonymouspapertrue
\else
  \anonymouspaperfalse
\fi

\ifanonymouspaper
  \usepackage[review]{acl}
\else
  \usepackage{acl}
\fi

\usepackage{times}
\usepackage{latexsym}
\usepackage[T1]{fontenc}
\usepackage[utf8]{inputenc}
\usepackage{microtype}
\usepackage{inconsolata}
\usepackage{url}
\usepackage{booktabs}
\usepackage{amsfonts}
\usepackage{amssymb}
\usepackage{nicefrac}
\usepackage{graphicx}
\usepackage{enumitem}

\usepackage{algorithm}
\usepackage{algpseudocode}

\usepackage{amsmath}
\usepackage{amsthm}

\theoremstyle{definition}

\usepackage{xcolor,colortbl}
\usepackage{listings}
\usepackage{multirow}
\usepackage{siunitx}[=v2]
\usepackage{cleveref}

\newcommand{\indistmark}{\textsuperscript{\(\spadesuit\)}}
\newcommand{\oodmark}{\textsuperscript{\(\heartsuit\)}}
\newcommand{\appendixexample}[1]{\par\medskip\noindent\textbf{#1}\par\smallskip}
\lstdefinestyle{appendixpython}{
  language=Python,
  basicstyle=\ttfamily\scriptsize,
  keywordstyle=\ttfamily\scriptsize,
  commentstyle=\ttfamily\scriptsize,
  stringstyle=\ttfamily\scriptsize,
  breaklines=true,
  breakatwhitespace=false,
  columns=fullflexible,
  keepspaces=true,
  showstringspaces=false,
  frame=single,
  framerule=0.2pt,
  rulecolor=\color{black!25},
  xleftmargin=0.2em,
  xrightmargin=0.2em,
  aboveskip=3pt,
  belowskip=3pt
}

\title{RACE: Scalable Statistical Estimation of Functional Consistency in LLM Neurons}

\ifanonymouspaper
  \author{}
\else
  \author{%
    \textbf{Runyu Wang}\textsuperscript{1} \quad
    \textbf{Bo Liu}\textsuperscript{2} \quad
    \textbf{Xiaxin Zhang}\textsuperscript{1} \quad
    \textbf{Yu Han}\textsuperscript{1} \quad
    \textbf{Jiawei Cao}\textsuperscript{1} \\
    \textbf{Xiaoye Zhang}\textsuperscript{3} \quad
    \textbf{Zhe Zhang}\textsuperscript{4} \quad
    \textbf{Yifan Yang}\textsuperscript{4} \quad
    \textbf{Peng Ping}\textsuperscript{1}\thanks{Corresponding author.} \\
    {\small \textsuperscript{1}School of Transportation and Civil Engineering, Nantong University} \\
    {\small \textsuperscript{2}Chongqing University of Post and Telecommunications} \\
    {\small \textsuperscript{3}China Southern Power Grid Company Limited \quad \textsuperscript{4}Meituan} \\
    {\scriptsize \texttt{2430310032@stmail.ntu.edu.cn} \quad \texttt{s250201066@stu.cqupt.edu.cn} \quad \texttt{2433320001@stmail.ntu.edu.cn} \quad \texttt{2433320018@stmail.ntu.edu.cn}} \\
    {\scriptsize \texttt{2233110297@stmail.ntu.edu.cn} \quad \texttt{xiaoyz@whu.edu.cn} \quad \texttt{zhangzhecnjs@gmail.com} \quad \texttt{yangyifan@meituan.com} \quad \texttt{pingpeng@ntu.edu.cn}}
  }
\fi

\begin{document}

\maketitle

\begin{abstract}
Discovering stable neuron behavior across entire domains remains a challenge in mechanistic interpretability.
Existing methods often rely on instance-level point estimates or computationally expensive procedures, which either obscure population-level variability or limit scalable domain-wide analysis.
We present \textbf{\textit{RACE}} (Residual Alignment for Consistency Estimation), a forward-pass statistical framework that evaluates the domain-wide functional consistency of Transformer neurons.
Compared with gradient-based point estimates, RACE produces neuron rankings that yield more domain-specific effects under perturbation.
Token-distribution shifts support the connection between the selected neurons and the target domain, while scoring requires roughly one-hundredth of the computational overhead of the gradient-based methods.
\ifanonymouspaper
\else
  Code is available at \url{https://github.com/Nexround/RACE}.
\fi

\end{abstract}

\section{Introduction}\label{sec:intro}

Recent advances in mechanistic interpretability have improved our understanding of Transformer-based Large Language Models (LLMs)~\citep{DBLP:journals/tist/ZhaoCYLDCWYD24,rai2024practical}. Three dominant approaches have emerged: causal tracing and knowledge localization techniques that identify critical model pathways~\citep{damai2021knowledge,kevin2022locating}, gradient-based attribution methods that quantify parameter importance~\citep{sundararajan2017axiomatic,achtibat2024attnlrp}, and sparse autoencoders that extract interpretable neuron activations~\citep{bricken2023monosemanticity,shu-etal-2025-survey}.

\begin{figure}[t]
  \centering
  \includegraphics[width=\columnwidth]{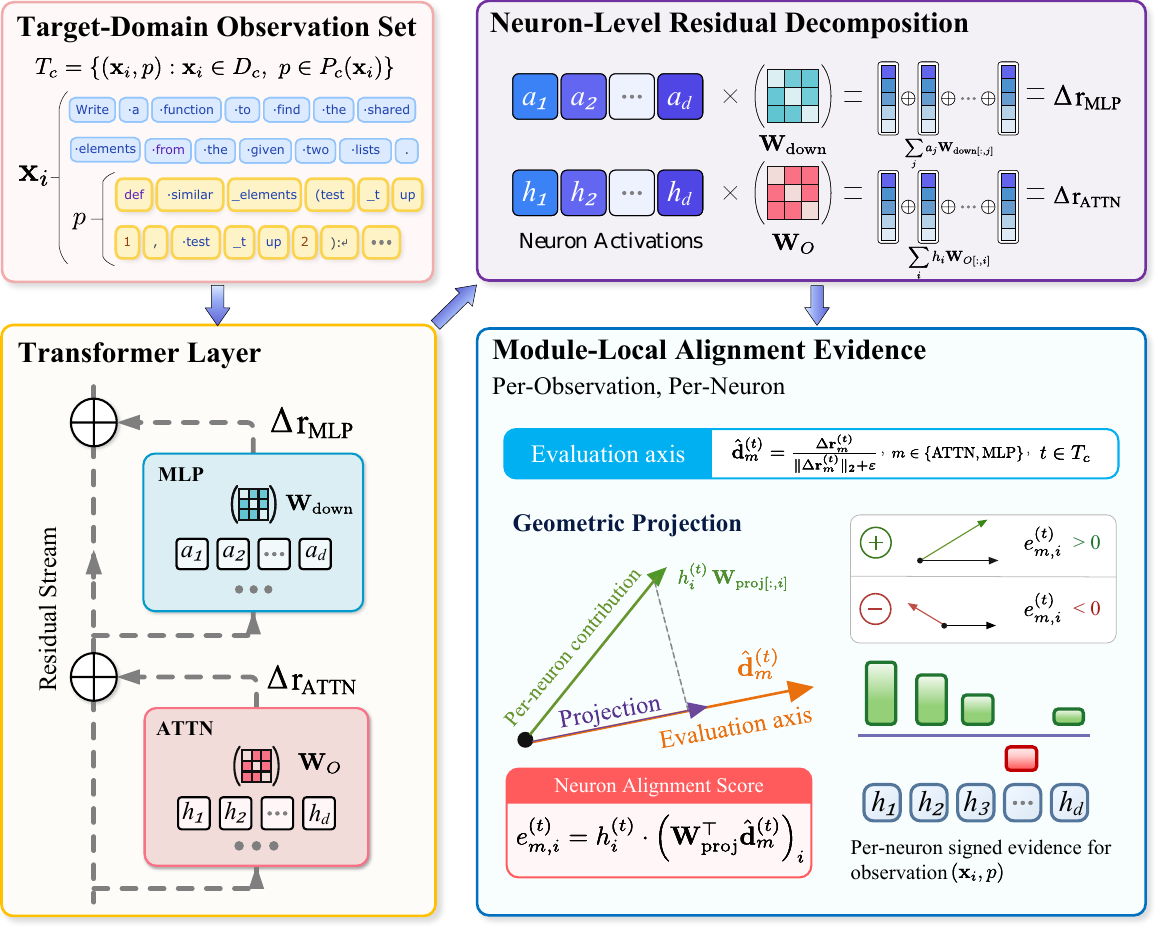}
  \caption{Overview of Residual-Direction Alignment (RDA), the per-observation evidence stage of RACE.
  RDA decomposes each module update into neuron writes and projects them onto the normalized update direction, yielding signed evidence \(e_{m,j}^{(t)}\).
  RACE aggregates this evidence across observations to estimate functional consistency.}
  \label{fig:race_rda}
\end{figure}

While these methods excel at providing instance-level explanations, they face a limitation: many real-world applications, including model capability auditing, domain-specific pruning, and behavioral steering, demand population-level characterizations of model components.
Specifically, these tasks require rankings of task-relevant neurons that remain consistent across diverse inputs.

Recent work has attempted to bridge this gap through task-neuron mapping strategies: causal gradient variation localizes neurons that selectively affect target tasks~\citep{song-etal-2024-large}, while gradient attribution links task-specific neuron overlap to cross-task generalization~\citep{leng-xiong-2025-towards}.
However, existing pipelines suffer from two constraints:
computational inefficiency stems from iterative gradient calculations and intervention operations, creating prohibitive overhead. Meanwhile, statistical oversimplification emerges when aggregating neuron contributions into task-level averages, which obscures sample-to-sample variation patterns.

To overcome these limitations, we draw upon two empirical observations: specific LLM capabilities rely on sparse subsets of neurons~\citep{frankle2019lottery, frantar2023sparsegpt}, and these neurons respond uniquely to semantically coherent inputs~\citep{voita-etal-2024-neurons, huang-etal-2025-neuron}.
Since neuron activations directly modulate the residual stream, we hypothesize that during the forward pass, neurons serving consistent domain-specific functions will deposit contribution distributions over this stream that differ from those of non-domain-specific neurons in a statistically significant manner.

Accordingly, we introduce \textbf{\textit{RACE}} (Residual Alignment for Consistency Estimation), a statistical estimation framework that scores every neuron at every layer for \emph{functional consistency} with respect to a \emph{target-domain observation set} (\S\ref{subsec:problem_formulation}).
RACE consists of two stages: (1)~decomposing each module's residual stream update into per-neuron contributions and evaluating their alignment with the module's output direction via Residual-Direction Alignment (RDA) at forward-pass cost; and (2)~distilling noisy per-observation signals into posterior distributions over each neuron's mean alignment and variance through Bayesian aggregation.

We evaluate RACE on 4B--32B LLMs across code generation, mathematical reasoning, and fine-grained behavioral control.
Targeted suppression shows RACE selectively disrupts target capabilities while preserving non-target behaviors.
Empirical results confirm RACE’s computational efficiency and demonstrate its superiority over gradient-based baselines.
Ablation studies validate the contribution of each RACE component.

\section{Method}\label{sec:method}

\subsection{Problem Formulation}\label{subsec:problem_formulation}

Let \(\mathcal{M}: \mathcal{X} \to \mathcal{Y}\) be a trained Transformer with layers \(l=1,\ldots,L\) and modules \(m \in \{\mathrm{ATTN}, \mathrm{MLP}\}\), denoting attention and multilayer perceptron modules, respectively.
We write \(u=(l,m,j)\) for a neuron in a fixed layer--module pair.

\noindent\textbf{Target-Domain Observation Set.} A \emph{target domain} \(c\) is specified by a predicate \(\phi_c: \mathcal{X} \to \{\mathrm{True}, \mathrm{False}\}\) that induces the input population \(\mathcal{D}_c = \{\mathbf{x} \in \mathcal{X} : \phi_c(\mathbf{x}) = \mathrm{True}\}\).
In practice, we approximate this with a finite sample \(D_c = \{\mathbf{x}_i\}_{i=1}^{N} \subseteq \mathcal{D}_c\).
Because the Transformer is token-position dependent, the auditing protocol evaluates one or more positions per input, yielding the operational observation set
\begin{equation}
    \begin{aligned}
        T_c &= \{(\mathbf{x}_i, p) : \mathbf{x}_i \in D_c,\; p \in \mathcal{P}_c(\mathbf{x}_i)\},\\
        n &= |T_c|,
    \end{aligned}
\end{equation}
where \(\mathcal{P}_c(\mathbf{x})\) denotes the analyzed token positions.

\noindent\textbf{Functional Consistency.} The \emph{functional consistency score} of neuron \(j\) with respect to \(D_c\) reflects how well its contribution distribution satisfies two desiderata.
Let \(e_j^{(t)} \in \mathbb{R}\) denote neuron \(j\)'s signed contribution for observation \(t \in T_c\), with the sign defined relative to the observation-specific normalized module-update direction \(\hat{\mathbf{d}}^{(t)}\) (Eq.~\ref{eq:eval_axis}).
The score evaluates:
\begin{itemize}
    \item \textbf{Magnitude:} \(\mathbb{E}_{t \in T_c}[e_j^{(t)}]\) is large and positive relative to other neurons in the same module.
    \item \textbf{Stability:} \(\text{Var}_{t \in T_c}[e_j^{(t)}]\) is small---the contribution is not driven by a few outlier observations.
\end{itemize}
Neurons that activate strongly on a handful of observations but are silent on most, or whose signed contributions change direction across observations, fail to satisfy these conditions despite potentially high average unsigned magnitude.
We denote by \(\mathcal{A}_{\mathcal{M}}(U_{l,m}, T_c) \in \mathbb{R}^{|U_{l,m}|}\) the vector of functional consistency scores for all neurons in module \(m\) at layer \(l\), given observation set \(T_c\).

To estimate \(\mathcal{A}_{\mathcal{M}}(U_{l,m}, T_c)\) across all layers and modules, RACE frames consistency evaluation as a problem of statistical inference over the evidence set \(E_{j,c}\triangleq\{e_j^{(t)}\}_{t\in T_c}\) (hereafter, we drop the layer and module subscripts for readability and simply write \(j\) for the audited neuron).
Specifically, we model each neuron as possessing latent evidence-distribution parameters \(\theta_j = (\mu_j, \sigma_j^2)\), treating its per-observation evidence as noisy realizations from an underlying distribution:
\begin{equation}
    e_j^{(t)} \mid \theta_j \sim P(\cdot \mid \mu_j, \sigma_j^2), \quad \forall\, t \in T_c \label{eq:generative}
\end{equation}
Functional consistency scoring then becomes a matter of posterior inference over the collected evidence:
\begin{equation}
    P(\theta_j \mid E_{j,c}) \propto \prod_{t \in T_c} P\!\left(e_j^{(t)} \mid \mu_j, \sigma_j^2\right) \cdot P(\mu_j, \sigma_j^2). \label{eq:posterior_objective}
\end{equation}
This posterior formulation directly conceptually maps to our two desiderata: the posterior over \(\mu_j\) captures the functional magnitude and direction, while the posterior over \(\sigma_j^2\) and the epistemic uncertainty in \(\mu_j\) jointly encode stability.

To estimate this posterior in practice, RACE employs a two-stage pipeline.
First, RDA computes the module-local evidence \(e_j^{(t)}\) from a single forward pass for each observation (\S\ref{subsec:rda}).
Second, Bayesian aggregation infers \(\theta_j\) from this evidence set (\S\ref{subsec:aggregation}) using a Normal-Inverse-Gamma (NIG) conjugate model.
This aggregation yields closed-form posterior mean estimates and variance-calibrated Conservative Alignment Magnitude (CAM) scores for neuron selection (\S\ref{subsec:uncertainty}).
For intervention settings that require disentangling target-specific neurons from broadly active ones, we introduce Reference-Set Filtering (RSF) as a filtration step (\S\ref{subsec:reference_set_filtering}).

\subsection{Residual-Direction Alignment}\label{subsec:rda}

RDA provides the per-observation evidence for Bayesian aggregation.
It projects each neuron's weighted output onto the normalized residual update of its host module, thereby avoiding raw-activation proxies~\citep{shrikumar2017learning, kevin2022locating} and the computational overhead of gradient-based attribution~\citep{sundararajan2017axiomatic}.

\subsubsection{Neuron-Level Residual Decomposition}\label{subsubsec:residual_decomp}
The residual stream \(\mathbf{r}\) flows from layer to layer, with each module contributing \(\Delta\mathbf{r}_{\text{module}}\): \(\mathbf{r}_{\text{out}} = \mathbf{r}_{\text{in}} + \Delta\mathbf{r}_{\text{module}}\).
For a fixed observation and layer, this update decomposes naturally into per-neuron contributions:
\begin{align}
    \Delta\mathbf{r}_{\text{ATTN}} &= \mathbf{W}_{\text{O}} \mathbf{h} = \sum_i h_i \mathbf{W}_{\text{O}_{[:,i]}} \label{eq:att_decomp}\\
    \Delta\mathbf{r}_{\text{MLP}} &= \mathbf{W}_{\text{down}} \mathbf{a} = \sum_j a_j \mathbf{W}_{\text{down}_{[:,j]}} \label{eq:mlp_decomp}
\end{align}
where \(\mathbf{h} \in \mathbb{R}^{d_{\text{model}}}\) is the concatenated attention head output and \(\mathbf{a} \in \mathbb{R}^{d_{\text{MLP}}}\) is the MLP intermediate activation before the down projection.
Each term is a rank-1 sub-update to the residual stream, scaled by activation magnitude.
This sub-update view follows prior work~\citep{geva2021transformer, geva2022transformer}.
RACE audits these same additive units: an MLP neuron \(j\) is the intermediate channel contributing \(a_j \mathbf{W}_{\text{down}_{[:,j]}}\), while an attention neuron denotes an output-channel contribution \(h_i \mathbf{W}_{\text{O}_{[:,i]}}\)~\citep{elhage2021mathematical, damai2021knowledge, yu2024neuron}.

\subsubsection{Module-Local Alignment Evidence}\label{subsubsec:alignment_scores}
RDA uses each module's output \(\Delta\mathbf{r}\) as the evaluation axis.
Appendix~\ref{sec:appendix_module_local_axis} discusses why this module-local choice is better matched to the evidence that RACE aims to collect.
For an observation \(t\) at layer \(l\), we define:
\begin{align}
    \hat{\mathbf{d}}_{\text{ATTN}}^{(t)} &= \frac{\Delta\mathbf{r}_{\text{ATTN}}^{(t)}}{\lVert\Delta\mathbf{r}_{\text{ATTN}}^{(t)}\rVert_2 + \varepsilon}\\
    \hat{\mathbf{d}}_{\text{MLP}}^{(t)} &= \frac{\Delta\mathbf{r}_{\text{MLP}}^{(t)}}{\lVert\Delta\mathbf{r}_{\text{MLP}}^{(t)}\rVert_2 + \varepsilon} \label{eq:eval_axis}
\end{align}
where \(\varepsilon\) is a small constant for numerical stability.
The unit vector \(\hat{\mathbf{d}}^{(t)}\) represents the direction of the module's additive update to the residual stream for observation \(t\).

Given this observation-specific axis, we first compute the directional alignment coefficient \(s_j^{(t)}\), which measures the geometric agreement between the neuron's output vector and the module update direction.
We then multiply this coefficient by the corresponding activation to obtain the activation-modulated signed evidence:
\begin{align}
    \mathbf{s}_{\text{ATTN}}^{(t)} &= \mathbf{W}_{\text{O}}^T \hat{\mathbf{d}}_{\text{ATTN}}^{(t)}, \quad e_{\text{ATTN},i}^{(t)} = h_i^{(t)} \cdot s_{\text{ATTN},i}^{(t)} \label{eq:att_contrib}\\
    \mathbf{s}_{\text{MLP}}^{(t)} &= \mathbf{W}_{\text{down}}^T \hat{\mathbf{d}}_{\text{MLP}}^{(t)}, \quad e_{\text{MLP},j}^{(t)} = a_j^{(t)} \cdot s_{\text{MLP},j}^{(t)} \label{eq:mlp_contrib}
\end{align}
The resulting scalar \(e_j^{(t)} \in \mathbb{R}\) is the per-observation evidence recorded by RDA.
Its sign records whether the neuron's weighted output is aligned with or opposed to the module's output direction for that observation.
Sign fluctuation across \(T_c\) is naturally handled by the Bayesian aggregation stage (\S\ref{subsec:aggregation}), where unstable evidence increases posterior uncertainty.
Because the module update is an exact sum of per-neuron writes, RDA
provides a signed module-local accounting of each neuron's  participation in the realized update. Appendix~\ref{sec:appendix_residual_alignment_justification} develops this argument in full.
Appendix~\ref{sec:appendix_evidence_strategies} details how the observation set is constructed and how per-observation evidence is logged for experiments.

\subsection{Evidence Aggregation}\label{subsec:aggregation}

We aggregate the noisy evidence \(\{e_j^{(t)}\}_{t \in T_c}\) over the observation set \(T_c\) induced by \(D_c\) using a NIG conjugate model.

\subsubsection{Modeling Functional Consistency}\label{subsubsec:nig_formulation}

For each neuron \(j\) in a given module at layer \(l\), we model the signed alignment scores as:
\begin{equation}
    e_j^{(t)} \sim \mathcal{N}(\mu_j, \sigma_j^2), \quad \forall\, t \in T_c \label{eq:nig_likelihood}
\end{equation}
Treating the Gaussian likelihood as a conjugate working model for the evidence mean and dispersion, we place a conjugate NIG prior \((\mu_j, \sigma_j^2) \sim \text{NIG}(\mu_0, \lambda_0, \alpha_0, \beta_0)\), where \(\sigma_j^2 \sim \text{Inv-Gamma}(\alpha_0, \beta_0)\) and \(\mu_j \mid \sigma_j^2 \sim \mathcal{N}(\mu_0, \sigma_j^2 / \lambda_0)\).

This conjugate specification yields analytic posterior updates for each neuron.

\subsubsection{Posterior Updates}\label{subsubsec:bayesian_update}

The closed-form update produces two quantities used by RACE scoring.
The first is a posterior mean \(\mu_{n,j}\) that estimates signed alignment strength.
The second is a posterior uncertainty term that penalizes noisy or scarce evidence.

\paragraph{Sufficient Statistics.} Given the \(n\) evidence values \(\{e_j^{(t)}\}_{t \in T_c}\) for neuron \(j\), the posterior update requires only the empirical mean and centred sum of squares:
\begin{equation}
    \bar{e}_j = \frac{1}{n}\sum_{t \in T_c} e_j^{(t)}, \quad \mathrm{SS}_j = \sum_{t \in T_c} \left(e_j^{(t)} - \bar{e}_j\right)^2, \label{eq:evidence_sum}
\end{equation}
from which the empirical standard deviation is \(\hat{\sigma}_j = \sqrt{\mathrm{SS}_j/(n-1)}\).

\paragraph{Posterior Update.} The NIG conjugacy yields a closed-form posterior with the same functional form:
\begin{equation}
    (\mu_j, \sigma_j^2) \mid \{e_j^{(t)}\}_{t \in T_c} \sim \text{NIG}(\mu_{n,j}, \lambda_n, \alpha_n, \beta_{n,j}) \label{eq:posterior_update}
\end{equation}
where the posterior hyperparameters are updated via simple arithmetic:
\begin{align}
    \lambda_n &= \lambda_0 + n, \label{eq:nig_lambda}\\[-6pt]
    \alpha_n &= \alpha_0 + \tfrac{n}{2}, \label{eq:nig_alpha}\\[-6pt]
    \mu_{n,j} &= \frac{\lambda_0 \mu_0 + n \bar{e}_j}{\lambda_n}, \label{eq:nig_mu}\\[-2pt]
    \beta_{n,j} &= \beta_0 + \tfrac{\mathrm{SS}_j}{2} + \frac{\lambda_0 n (\bar{e}_j - \mu_0)^2}{2\lambda_n}. \label{eq:nig_beta}
\end{align}
Maintaining these sufficient statistics costs \(O(1)\) per neuron per observation.
The posterior parameters are then obtained in closed form.

\paragraph{Posterior Mean Evidence.} The group-level score is the signed posterior mean, directly instantiating Eq.~\eqref{eq:posterior_objective}:
\begin{equation}
    \Phi_c(j) = \mu_{n,j} = \frac{\lambda_0 \mu_0 + n \bar{e}_j}{\lambda_0 + n} \label{eq:importance_score}
\end{equation}
The sign records the dominant alignment direction of the evidence, while values near zero indicate weak or inconsistent average alignment.
The next subsection converts this posterior mean and its marginal uncertainty into CAM.

\subsection{Uncertainty Quantification}\label{subsec:uncertainty}

\noindent\textbf{Posterior Uncertainty over Mean Evidence.} Integrating out \(\sigma_j^2\), the marginal posterior for the mean evidence follows a Student's \(t\)-distribution:
\begin{equation}
    \begin{aligned}
        \mu_j \mid \{e_j^{(t)}\}_{t \in T_c}
        &\sim t_{2\alpha_n}\!\left(\mu_{n,j},\; \sigma_{\mu,j}\right),\\
        \sigma_{\mu,j}^2
        &= \frac{\beta_{n,j}}{\alpha_n \lambda_n}
    \end{aligned}
    \label{eq:t_marginal}
\end{equation}
where the second argument is the Student's \(t\) scale parameter.
This scale \(\sigma_{\mu,j}\) quantifies uncertainty regarding the average signed contribution, producing wider intervals for scarce or noisy evidence.

\noindent\textbf{Variance Sources.} The same \(\beta_{n,j}\) term also determines the posterior expected variance of the alignment evidence across observations, \(\mathbb{E}[\sigma_j^2 \mid \{e_j^{(t)}\}_{t \in T_c}] = \beta_{n,j} / (\alpha_n - 1)\) for \(\alpha_n > 1\).
Its data-driven component \(\mathrm{SS}_j/2\) captures cross-observation variability in the alignment evidence.
The prior-data term \(\lambda_0 n(\bar{e}_j - \mu_0)^2/(2\lambda_n)\) regularizes evidence relative to the neutral prior \(\mu_0=0\).
Thus bursty, noisy, or sign-fluctuating evidence inflates \(\beta_{n,j}\), increasing \(\sigma_{\mu,j}\) and reducing confidence in the neuron's mean alignment strength.
As \(n\) grows, \(\lambda_n\) and \(\alpha_n\) increase, shrinking the posterior uncertainty over \(\mu_j\) when the evidence remains stable.

\noindent\textbf{Conservative Alignment Magnitude.} We define a scoring criterion that provides a one-sided conservative lower credible bound on each neuron's positive module-output alignment.
The default RACE score is:
\begin{equation}
    \begin{aligned}
    \rho_j(\gamma)
    &= \mu_{n,j} - t_{1-\gamma,\,2\alpha_n}\sigma_{\mu,j},\\
    \text{CAM}_j(\gamma)
    &= \mathbb{I}[\mu_{n,j}>0]\,\max\!\left(0,\rho_j(\gamma)\right)
    \end{aligned}
    \label{eq:cam}
\end{equation}
where \(t_{1-\gamma,\,2\alpha_n}\) is the \((1{-}\gamma)\)-quantile of the Student's \(t\) with \(2\alpha_n\) degrees of freedom.
Formally, \(\text{CAM}_j(\gamma)\) is the lower \((1-\gamma)\)-credible bound on the positive posterior mean evidence under the marginal Student's \(t\)-posterior in Eq.~\eqref{eq:t_marginal}.
Neurons with large positive mean evidence and small posterior uncertainty receive high CAM scores, whereas neurons with few observations, unstable alignment, or negative posterior mean evidence are excluded or penalized through the confidence radius.
Throughout the paper, \textbf{RACE} denotes the default CAM ranking.
As an ablation, we also report \(\text{Neg. CAM}_j(\gamma)=\mathbb{I}[\mu_{n,j}<0]\max(0,-\mu_{n,j}-t_{1-\gamma,\,2\alpha_n}\sigma_{\mu,j})\).

\subsection{Reference-Set Filtering}\label{subsec:reference_set_filtering}

Because features represented in superposition can make individual neurons polysemantic~\citep{elhage2022superposition,scherlis2022polysemanticity,templeton2024scaling}, unfiltered target rankings may include neurons that support broad capabilities rather than neurons whose behavior is specific only to the target domain.
To disentangle domain-specific behavior from general capabilities during targeted interventions, we introduce RSF as an explicit filtering step.
We denote direct RACE scoring on dataset \(D\) by \(\mathbf{R}_{D}\), and reference-set filtered scoring by \(\mathbf{R}_{D_{\mathrm{tar}} \setminus D_{\mathrm{ref}}}\).
For each layer \(\ell\) and module type, let \(U^\ell\) be the layer-local neuron universe, \(B_{\mathrm{ref}}^\ell=\operatorname{Top}_{K_{\mathrm{ref}}}^{s_{\mathrm{ref}}}(U^\ell)\) be a reference exclusion set, and \(\Pi_{\mathrm{tar}}^\ell\) be the full target-domain ranking.
Here \(K_{\mathrm{ref}}\in\mathbb{N}_{>0}\) is the number of reference-selected neurons to exclude.
When the reference budget is instead specified as a fraction \(\tau_{\mathrm{ref}}\in(0,1]\), \(\operatorname{Top}_{\tau_{\mathrm{ref}}}^{s_{\mathrm{ref}}}(U^\ell)\) denotes the corresponding top-fraction exclusion set.
Let \(K_{\mathrm{sel}}\in\mathbb{N}_{>0}\) denote the number of target neurons retained for intervention in each layer and module.
RSF selects
\begin{equation}
    \begin{aligned}
    \Pi_{\mathrm{RSF}}^\ell
    &=
    \left[
    j \in \Pi_{\mathrm{tar}}^\ell
    :
    j \notin B_{\mathrm{ref}}^\ell
    \right],
    \\
    S_{\mathrm{spec}}^\ell
    &=
    \operatorname{First}_{K_{\mathrm{sel}}}\!\left(\Pi_{\mathrm{RSF}}^\ell\right).
    \end{aligned}
    \label{eq:reference_filtered_set}
\end{equation}
Operationally, RSF traverses the target ranking, skips reference-selected neurons, and stops when \(K_{\mathrm{sel}}\) neurons are selected; this preserves exactly \(K_{\mathrm{sel}}\) neurons per layer whenever \(|U^\ell \setminus B_{\mathrm{ref}}^\ell| \ge K_{\mathrm{sel}}\).
Appendix~\ref{sec:appendix_cross_domain_consistency} provides a cross-domain overlap analysis.

\section{Experiments}\label{sec:experiments}

We evaluate RACE across multiple domains and models.
Our experiments address two main questions: (1) \textbf{Efficacy}: Do RACE-selected neurons cause a disproportionate performance drop on target domains versus non-target domains when suppressed? (2) \textbf{Ablation}: Does Bayesian uncertainty quantification (CAM) outperform deterministic scoring heuristics?

\subsection{Experimental Setup}\label{subsec:setup}

\noindent\textbf{Evaluated Models.} We evaluate on Qwen3-4B-it~\citep{qwen3} (\texttt{Qwen3-4B-it-2507}), OLMo-3.1-32B-it~\citep{olmo2025olmo3}, and Llama-3.1-8B-it~\citep{llama31} (Appendix~\ref{sec:appendix_math_reference_filtered}); full model details are in Appendix~\ref{sec:appendix_eval_details}.

\noindent\textbf{Domain \& Benchmark Settings.} Each domain is instantiated by a scoring set for RACE-based neuron selection, a same-domain out-of-distribution (OOD) benchmark for validation, and non-target benchmarks for retention.
\begin{itemize}
    \item \textbf{Code:} MBPP+~\citep{austin2021program,evalplus} for scoring and HumanEval+~\citep{chen2021codex,evalplus} for OOD validation.
    \item \textbf{Math:} MATH-500~\citep{hendrycks2021math} for scoring and AMC~\citep{amc2023} for OOD validation.
    \item \textbf{Fine-grained Behavioral:} We construct \texttt{PyComp-1K}, a set of 1,000 AST-verified Python comprehension-containing statements from \texttt{bigcode/the-stack}~\citep{kocetkov2022stack,bigcode_the_stack_docs}, as a narrow code-behavior scoring domain (Appendix~\ref{sec:appendix_py_comp_dataset}).
\end{itemize}
Detailed benchmark evaluation setup, including evaluator configurations and scoring definitions, is provided in Appendix~\ref{sec:appendix_eval_details}.
Observation-collection and evidence-logging protocols for the evaluated corpora are summarized in Appendix~\ref{sec:appendix_evidence_strategies}.

\noindent\textbf{RACE Hyperparameters.} Unless otherwise specified, all experiments use the same RACE hyperparameter settings: \(\mu_0 = 0\), \(\lambda_0 = 1\), \(\alpha_0 = 1\), \(\beta_0 = 1\), and \(\gamma = 0.05\). We further discuss the robustness to prior settings and confidence levels in Appendix~\ref{sec:appendix_prior_settings}.

\noindent\textbf{Baselines \& Ablations.} Table~\ref{tab:baseline_ablation_definitions} reports the scoring formula used by each baseline or ablation.
We consider two groups of methods: external baselines and internal RACE ablations.
For external baselines, GxAct~\citep{kokhlikyan2020captum} and AttnLRP~\citep{achtibat2024attnlrp} serve as typical gradient attribution methods over the same \(T_c\), with the attribution target set to the benchmark answer token. Act. Mean serves as an activation-only control.
To isolate the effect of Bayesian uncertainty modeling, we introduce three internal ablations operating on the same evidence \(\{e_j^{(t)}\}_{t \in T_c}\) as RACE. Emp. Mean removes both the uncertainty penalty and prior regularization, thereby reducing to the raw empirical average. Emp. SNR replaces the Bayesian penalty with a frequentist variance penalty computed via \(\hat{\sigma}_j = \sqrt{\mathrm{SS}_j/(n-1)}\), and Neg. CAM acts as a negative-direction ablation.
Regarding the evaluation protocol, instance-level scores are lifted to domain-level neuron rankings by averaging positive neuron-level contributions over the induced token-position observation set \(T_c\), matching RACE's evidence aggregation granularity.
All methods adopt identical per-layer/per-module selection budgets and suppression protocols.
Under RSF, target and reference rankings are computed using the same method-specific scoring rule.

\begin{table}[t]
\centering
\scriptsize
\setlength{\tabcolsep}{3pt}
\begin{tabular}{p{0.18\columnwidth}p{0.76\columnwidth}}
\toprule
\textbf{Method} & \textbf{Neuron score \(S_j\)} \\
\midrule
GxAct & \(1/n\sum_{t\in T_c}\max(0,\mathrm{GxAct}_j^{(t)})\) \\
AttnLRP & \(1/n\sum_{t\in T_c}\max(0,\mathrm{LRP}_j^{(t)})\) \\
Act. Mean & \(1/n\sum_{t\in T_c}|a_j^{(t)}|\) \\
\midrule
Neg. CAM & \(\mathbb{I}[\mu_{n,j}<0]\max(0,-\mu_{n,j}-t_{1-\gamma,2\alpha_n}\sigma_{\mu,j})\) \\
Emp. Mean & \(\bar{e}_j\) \\
Emp. SNR & \(\max(0,\bar{e}_j) / (\hat{\sigma}_j+\varepsilon)\) \\
\bottomrule
\end{tabular}
\caption{
Baselines and ablations.
Each row gives the neuron score \(S_j\) used for ranking neuron \(j\) from \(n\) observations.
\(\mathrm{GxAct}_j^{(t)}\) and \(\mathrm{LRP}_j^{(t)}\) denote per-observation attribution scores for the two external baselines.}\label{tab:baseline_ablation_definitions}
\end{table}

\subsection{Depth-Wise Organization of CAM Scores}\label{subsec:race_visualization}

\begin{figure}[ht]
\centering
\includegraphics[width=\columnwidth]{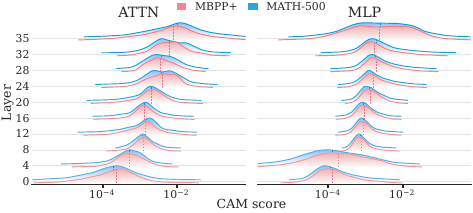}
\caption{
Layer-wise kernel density estimate (KDE) ridgeline distributions of RACE results.
On Qwen3-4B-it, ridges under \(\mathbf{R}_{\text{MBPP+}}\) and \(\mathbf{R}_{\text{MATH-500}}\) visualize the KDE of CAM scores for neurons in each module across selected layers.}\label{fig:cam_kde_ridge}
\end{figure}

Figure~\ref{fig:cam_kde_ridge} shows how CAM scores vary across layers and modules.
The ridgelines reveal a progressive rightward shift in positive CAM density with increasing depth, showing that high CAM scores concentrate in later layers.
Compared to the shallow and deep layers, the middle layers exhibit a sparser distribution of high-scoring neurons, especially in MLP modules.
The same depth-wise pattern appears on both MBPP+ and MATH-500, suggesting that it reflects a property shared across the two domains.

Consistent with prior analyses on Transformers' functionally stratified computation~\citep{ian2019bert,jawahar2019what,geva2022transformer,geva2023dissecting}, this observation implies that CAM scores are not directly comparable across layers.
A globally sorted list would be dominated by late-layer neurons, conflating alignment magnitude with network depth.
We therefore adopt a stratified selection protocol in the following interventions: neurons are ranked by CAM within each layer and module, and a fixed budget is allocated to every layer.
This per-layer selection rule preserves coverage over the model's hierarchical computation.

\subsection{Effectiveness Validation}\label{subsec:neuron_identification}

We validate the intervention relevance of the selected neurons through targeted suppression: during inference, we set their corresponding activation values to zero and measure the resulting performance changes.

To jointly quantify target-domain suppression effectiveness and general selectivity in a single scalar, we define the \textbf{Intervention Specificity Index (ISI)}:
\begin{equation}
    \text{ISI} =
    \max\left\{0, \log\left(\frac{\Delta_T}{\Delta_G + \delta}\right)\right\},
    \quad \delta = 0.01.
    \label{eq:isi}
\end{equation}
Here \(\Delta_T\) and \(\Delta_G\) are the average nonnegative relative accuracy drops over the target-domain (including OOD benchmarks) and general (non-target) benchmarks respectively, computed from the per-benchmark drop \(\Delta = \max\{0,(\text{Acc}_{\text{before}} - \text{Acc}_{\text{after}}) / \text{Acc}_{\text{before}}\}\).

The constant \(\delta\) is used for smoothing.
Higher ISI indicates stronger target-domain suppression with minimal non-target degradation.

\subsubsection{Domain-Specific Intervention}\label{subsubsec:llm_suppression}

\begin{figure*}[!t]
  \centering
  \includegraphics[width=\linewidth]{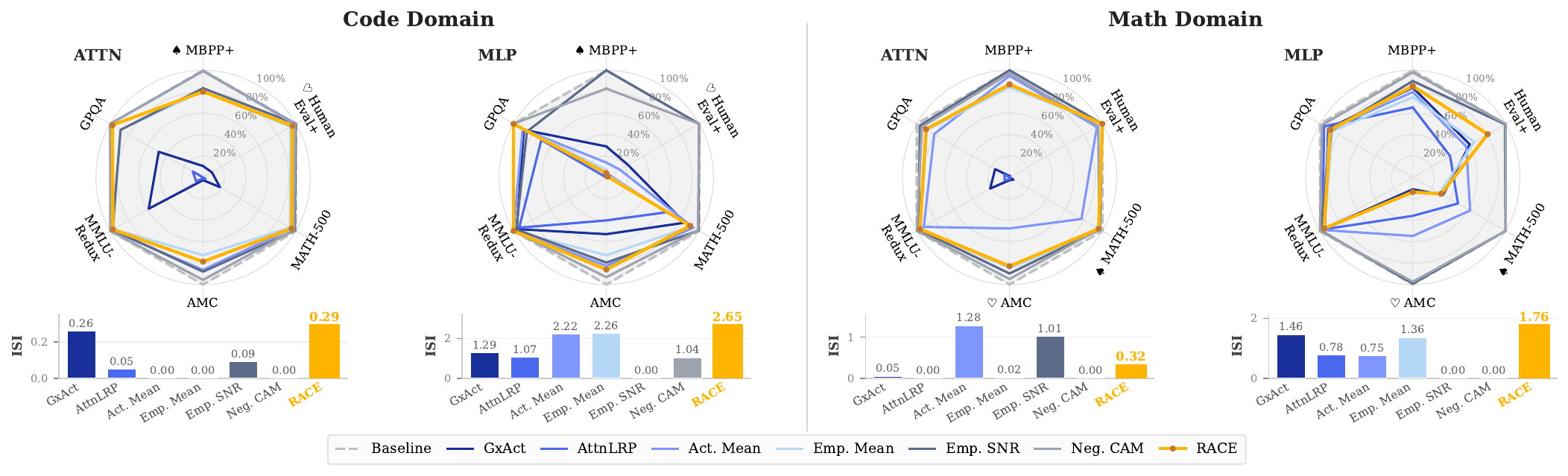}
  \caption{
  Suppression under the RSF strategies \(\mathbf{R}_{\text{MBPP+} \setminus \text{WikiText-2}}\) for the code domain and \(\mathbf{R}_{\text{MATH-500} \setminus \text{WikiText-2}}\) for the math domain on Qwen3-4B-it.
  All interventions suppress the top-\(1\%\) target-selected neurons within the corresponding module at each layer.
  Radar plots report post-suppression benchmark accuracy as a percentage of the original model baseline, separately for ATTN and MLP interventions.
  Bars below each radar report the corresponding ISI.}
  \label{fig:radar_code_math_suppression}
\end{figure*}

We evaluate two intervention strategies.
The \textbf{Vanilla Selection} strategy relies solely on the CAM scores computed on the target domain.
In contrast, the \textbf{RSF} strategy (\S\ref{subsec:reference_set_filtering}) explicitly controls for broadly shared language capabilities by filtering the candidate set against a general reference corpus.

\paragraph{Vanilla Selection.} Table~\ref{tab:code_suppression} reports the Code-domain results; the corresponding Qwen3-4B-it Math-domain results are provided in Appendix Table~\ref{tab:appendix_math_suppression_qwen3}.
We find that neurons selected under this setting cause severe overall generation degradation once the suppression budget reaches \(K_{\mathrm{sel}}{\geq}10\) per layer.
We thus report \(K_{\mathrm{sel}}{=}5\) to observe distinct effects.

\begin{table}[ht]
\centering
\setlength{\tabcolsep}{3pt}
\resizebox{\linewidth}{!}{
\begin{tabular}{c@{\hspace{3pt}}l@{\hspace{3pt}}S[table-format=2.2]@{\hspace{5pt}}S[table-format=2.2]@{\hspace{5pt}}S[table-format=2.2]@{\hspace{5pt}}S[table-format=2.2]@{\hspace{5pt}}S[table-format=2.2]@{\hspace{5pt}}S[table-format=2.2]@{\hspace{5pt}}S[table-format=1.2]}
\toprule
\textbf{Module} & \textbf{Method} & {\textbf{MBPP+\indistmark}} & {\textbf{HumanEval+\oodmark}} & {\textbf{MATH-500}} & {\textbf{AMC}} & {\textbf{MMLU-Redux}} & {\textbf{GPQA}} & {\textbf{ISI}\(\uparrow\)} \\
\midrule
\multirow{7}{*}{ATTN}
 & GxAct & 57.94 & 49.39 & 76.35 & 42.54 & 77.79 & 44.95 & 0.57 \\
 & AttnLRP & 47.09 & 43.90 & 90.25 & 77.61 & 76.79 & 45.96 & 2.00 \\
 & Act. Mean & 80.69 & 46.34 & 93.89 & 81.34 & 80.82 & 41.92 & 1.54 \\
 & Emp. Mean & 78.31 & 51.83 & 93.24 & 84.33 & 80.47 & 39.97 & 1.37 \\
 & Emp. SNR & 74.34 & 50.00 & 93.13 & 84.33 & 81.14 & 41.41 & 1.71 \\
 & Neg. CAM & 78.84 & 83.54 & 93.77 & 83.58 & 80.53 & 41.92 & 0.00 \\
\rowcolor[RGB]{236,244,252}\cellcolor{white} & \textbf{RACE} & 55.03 & 45.73 & 89.00 & 72.39 & 78.91 & 44.95 & 1.62 \\
\midrule
\multirow{7}{*}{MLP}
 & GxAct & 10.32 & 5.49 & 1.80 & 0.75 & 15.25 & 15.15 & 0.04 \\
 & AttnLRP & 74.87 & 83.54 & 95.04 & 85.82 & 81.84 & 49.49 & 1.15 \\
 & Act. Mean & 81.75 & 3.66 & 93.42 & 81.34 & 81.18 & 41.47 & 2.22 \\
 & Emp. Mean & 75.13 & 0.00 & 89.86 & 87.31 & 81.33 & 43.43 & 2.79 \\
 & Emp. SNR & 80.95 & 86.59 & 93.62 & 82.09 & 81.23 & 45.96 & 0.00 \\
 & Neg. CAM & 75.13 & 85.37 & 91.81 & 84.33 & 81.58 & 45.96 & 0.54 \\
\rowcolor[RGB]{236,244,252}\cellcolor{white} & \textbf{RACE} & 74.34 & 0.00 & 92.80 & 82.83 & 81.79 & 44.95 & 2.91 \\
\midrule
\multicolumn{2}{l}{Qwen3-4B-it} & 82.28 & 83.54 & 94.40 & 87.31 & 81.37 & 45.45 & {---} \\
\bottomrule
\end{tabular}
}
\caption{
Code domain with \(\mathbf{R}_{\text{MBPP+}}\): Benchmark Acc. (\%) and ISI on Qwen3-4B-it after suppressing the top \(K_{\mathrm{sel}}{=}5\) neurons per layer.
Superscript \indistmark{} marks the target domain \(D\), and \oodmark{} marks the same-domain OOD benchmark.
All reported scores are averaged over three runs; subsequent tables follow the same setting unless specified.
}\label{tab:code_suppression}
\end{table}

\paragraph{RSF Selection.} On Qwen3-4B-it, Figure~\ref{fig:radar_code_math_suppression} summarizes the RSF intervention results for both the code and math domains.
The corresponding tabular details are reported in Appendix Tables~\ref{tab:appendix_code_suppression_reference_filtered_qwen3} and~\ref{tab:appendix_math_suppression_reference_filtered_qwen3}.
To demonstrate that RACE's effectiveness scales to larger models, we further evaluate \(\mathbf{R}_{\text{MATH-500} \setminus \text{WikiText-2}}\) on OLMo-3.1-32B-it.

\begin{table}[ht]
\centering
\scriptsize
\setlength{\tabcolsep}{2pt}
\resizebox{\linewidth}{!}{
\begin{tabular}{c@{\hspace{3pt}}l@{\hspace{3pt}}S[table-format=2.2]@{\hspace{5pt}}S[table-format=2.2]@{\hspace{5pt}}S[table-format=2.2]@{\hspace{5pt}}S[table-format=2.2]@{\hspace{5pt}}S[table-format=2.2]@{\hspace{5pt}}S[table-format=2.2]@{\hspace{5pt}}S[table-format=2.2]@{\hspace{5pt}}S[table-format=2.2]@{\hspace{5pt}}S[table-format=1.2]@{\hspace{5pt}}S[table-format=1.2]}
\toprule
& & \multicolumn{2}{c}{\textbf{MATH-500\indistmark}} & \multicolumn{2}{c}{\textbf{AMC\oodmark}} & \multicolumn{2}{c}{\textbf{GPQA}} & \multicolumn{2}{c}{\textbf{MMLU-Redux}} & \multicolumn{2}{c}{\textbf{ISI}\(\uparrow\)} \\
\cmidrule(lr){3-4} \cmidrule(lr){5-6} \cmidrule(lr){7-8} \cmidrule(lr){9-10} \cmidrule(lr){11-12}
\textbf{Module} & \textbf{Method} & {\(1\%\)} & {\(5\%\)} & {\(1\%\)} & {\(5\%\)} & {\(1\%\)} & {\(5\%\)} & {\(1\%\)} & {\(5\%\)} & {\(1\%\)} & {\(5\%\)} \\
\midrule
\multirow{5}{*}{ATTN}
 & Act. Mean & 88.32 & 84.85 & 76.12 & 71.64 & 38.10 & 35.99 & 79.91 & 73.18 & 0.00 & 0.00 \\
 & Emp. Mean & 85.01 & 87.84 & 70.93 & 68.66 & 45.03 & 43.52 & 81.54 & 77.12 & 0.00 & 0.00 \\
 & Emp. SNR & 86.41 & 86.25 & 71.64 & 67.91 & 44.02 & 44.54 & 83.47 & 79.95 & 0.00 & 0.00 \\
 & Neg. CAM & 85.97 & 86.62 & 77.61 & 73.13 & 47.56 & 45.53 & 84.01 & 84.67 & 0.00 & 0.01 \\
\rowcolor[RGB]{236,244,252}\cellcolor{white} & \textbf{RACE} & 85.61 & 84.43 & 76.86 & 65.67 & 47.74 & 42.49 & 82.63 & 80.35 & 0.00 & 0.00 \\
\midrule
\multirow{5}{*}{MLP}
 & Act. Mean & 63.87 & 14.71 & 21.64 & 10.45 & 40.40 & 36.36 & 81.32 & 75.65 & 1.47 & 1.50 \\
 & Emp. Mean & 73.95 & 21.43 & 25.37 & 13.43 & 41.41 & 38.89 & 80.81 & 73.46 & 1.35 & 1.50 \\
 & Emp. SNR & 76.89 & 44.54 & 23.88 & 18.66 & 23.74 & 21.72 & 52.81 & 46.21 & 0.00 & 0.20 \\
 & Neg. CAM & 92.40 & 97.48 & 79.85 & 79.85 & 52.02 & 50.51 & 84.32 & 85.16 & 0.00 & 0.00 \\
\rowcolor[RGB]{236,244,252}\cellcolor{white} & \textbf{RACE} & 56.80 & 6.72 & 14.18 & 7.46 & 39.83 & 37.79 & 82.72 & 77.79 & 1.65 & 1.73 \\
\midrule
\multicolumn{2}{l}{OLMo-3.1-32B-it} & \multicolumn{2}{c}{86.40} & \multicolumn{2}{c}{79.85} & \multicolumn{2}{c}{48.6} & \multicolumn{2}{c}{84.70} & \multicolumn{2}{c}{---} \\
\bottomrule
\end{tabular}
}
\caption{
Math domain with \(\mathbf{R}_{\text{MATH-500} \setminus \text{WikiText-2}}\): Benchmark Acc. (\%) and ISI on OLMo-3.1-32B-it after suppressing top-\(1\%\) and top-\(5\%\) neurons per layer.
}\label{tab:math_suppression}
\end{table}

\noindent\textbf{Observation:} RACE generally yields stronger target-domain selectivity than gradient-based methods, supporting the use of RDA as a gradient-free evidence source. The pronounced OOD degradation suggests RACE scores are an effective target-domain proxy.
Increasing the suppression budget produces larger target-domain effects, while RSF limits degradation on the non-target benchmarks.

Figure~\ref{fig:radar_code_math_suppression} shows this specificity holds for both code and math: RACE-selected suppression under RSF hurts target-domain performance more than non-target performance. However, effects vary by module: MLP interventions cause significant same-domain OOD drops, while attention interventions show weaker domain-specific degradation.

This asymmetry is clearer when comparing vanilla selection to RSF: Under vanilla selection, ATTN interventions produce marked target drops even at low budgets, but RSF's ISI for ATTN remains substantially lower than for MLP, especially on the 32B model. We attribute this to ATTN's high-scoring neurons being sparse yet broadly reusable: RSF removes these general-purpose neurons via reference filtering, leaving a weaker domain-specific signal. This aligns with known sparsity in ATTN's $\mathbf{W}_O$~\citep{michel2019sixteen}, suggesting ATTN's target-domain high-activation neurons also serve broader linguistic functions.

\subsubsection{Distributional Verification}\label{subsubsec:causal_disruption}

\begin{table}[ht]

\centering

\scriptsize
\setlength{\tabcolsep}{3pt}
\resizebox{\linewidth}{!}{
\begin{tabular}{lc@{\hspace{6pt}} c c c @{\hspace{12pt}} c c c}

\toprule
& & \multicolumn{3}{c}{\textbf{(a) }\(\mathbf{R}_{\text{MBPP+} \setminus \text{WikiText-2}}\)} & \multicolumn{3}{c}{\textbf{(b) }\(\mathbf{R}_{\text{MATH-500} \setminus \text{WikiText-2}}\)} \\
\cmidrule(lr){3-5} \cmidrule(lr){6-8}
\textbf{Metric} & \textbf{Module} & {\textbf{MBPP+\indistmark}} & {\textbf{MATH-500}} & {\textbf{WikiText-2}} & {\textbf{MBPP+}} & {\textbf{MATH-500\indistmark}} & {\textbf{WikiText-2}} \\
\midrule
\multirow{2}{*}{\(\Delta_\mathrm{PPL}\)}
 & ATTN & \multicolumn{1}{S[table-format=+3.2,retain-explicit-plus,table-space-text-post={\%}]}{+6.17\%} & \multicolumn{1}{S[table-format=+3.2,retain-explicit-plus,table-space-text-post={\%}]}{+6.18\%} & \multicolumn{1}{S[table-format=+3.2,retain-explicit-plus,table-space-text-post={\%}]}{-2.49\%} & \multicolumn{1}{S[table-format=+3.2,retain-explicit-plus,table-space-text-post={\%}]}{+4.52\%} & \multicolumn{1}{S[table-format=+3.2,retain-explicit-plus,table-space-text-post={\%}]}{+7.29\%} & \multicolumn{1}{S[table-format=+3.2,retain-explicit-plus,table-space-text-post={\%}]}{-2.74\%} \\
 & MLP  & \multicolumn{1}{S[table-format=+3.2,retain-explicit-plus,table-space-text-post={\%}]}{+77.31\%} & \multicolumn{1}{S[table-format=+3.2,retain-explicit-plus,table-space-text-post={\%}]}{+23.98\%} & \multicolumn{1}{S[table-format=+3.2,retain-explicit-plus,table-space-text-post={\%}]}{+2.76\%} & \multicolumn{1}{S[table-format=+3.2,retain-explicit-plus,table-space-text-post={\%}]}{+25.41\%} & \multicolumn{1}{S[table-format=+3.2,retain-explicit-plus,table-space-text-post={\%}]}{+129.04\%} & \multicolumn{1}{S[table-format=+3.2,retain-explicit-plus,table-space-text-post={\%}]}{+2.60\%} \\
\midrule
\multirow{2}{*}{\(\bar{D}_{\mathrm{KL}}\)}
 & ATTN & \multicolumn{1}{S[table-format=1.2]}{0.10} & \multicolumn{1}{S[table-format=1.2]}{0.08} & \multicolumn{1}{S[table-format=1.2]}{0.03} & \multicolumn{1}{S[table-format=1.2]}{0.09} & \multicolumn{1}{S[table-format=1.2]}{0.09} & \multicolumn{1}{S[table-format=1.2]}{0.03} \\
 & MLP  & \multicolumn{1}{S[table-format=1.2]}{0.58} & \multicolumn{1}{S[table-format=1.2]}{0.22} & \multicolumn{1}{S[table-format=1.2]}{0.02} & \multicolumn{1}{S[table-format=1.2]}{0.22} & \multicolumn{1}{S[table-format=1.2]}{0.86} & \multicolumn{1}{S[table-format=1.2]}{0.03} \\
\bottomrule
\end{tabular}
}
\caption{Distributional disruption on Qwen3-4B-it under \(\mathbf{R}_{D_{\mathrm{tar}} \setminus \text{WikiText-2}}\), after suppressing the top-1\% neurons within the selected module at each layer.}\label{tab:ppl_kl}
\end{table}

Beyond coarse accuracy drops, we examine the mechanism of disruption at the token-distribution level using relative perplexity degradation \(\Delta_\mathrm{PPL}\) (Eq.~\ref{eq:delta_ppl}) and mean forward Kullback--Leibler divergence \(\bar{D}_{\mathrm{KL}}\) (Eq.~\ref{eq:mean_kl}).

The distributional metrics (Table~\ref{tab:ppl_kl}) reveal a drastic asymmetric disruption, particularly for MLP interventions.
Suppressing RACE-selected MLP neurons triggers severe target-domain distribution collapse (\(\Delta_\mathrm{PPL}\) reaching 77.31\% on MBPP+ and 129.04\% on MATH-500, alongside massive \(\bar{D}_{\mathrm{KL}}\) shifts), while leaving the reference corpus (WikiText-2) almost entirely unperturbed.
This targeted disruption indicates that RACE decouples and localizes function-specific neurons without collapsing the model's underlying linguistic competence.
As a byproduct, ATTN interventions exhibit significantly weaker distributional contrasts, suggesting lower functional sparsity and weaker sensitivity to perturbations compared to the MLP module.
Appendix~\ref{sec:appendix_no_rsf_distributional_qwen3} reports w/o-RSF distributional disruption results.

\subsubsection{Fine-Grained Behavioral Steering}\label{subsubsec:comprehension_suppression}

\begin{table}[ht]
\centering
\scriptsize
\setlength{\tabcolsep}{6pt}
\resizebox{\columnwidth}{!}{
\begin{tabular}{l@{\hspace{6pt}}S[table-format=2.0]S[table-format=2.0]S[table-format=2.2]@{\hspace{12pt}}S[table-format=2.0]S[table-format=2.0]S[table-format=2.2]}
\toprule
& \multicolumn{3}{c}{\textbf{MBPP+}} & \multicolumn{3}{c}{\textbf{HumanEval+}} \\
\cmidrule(lr){2-4} \cmidrule(lr){5-7}
\textbf{Strategy} & {\textbf{Records}} & {\textbf{Total}} & {\textbf{Pass\%}} & {\textbf{Records}} & {\textbf{Total}} & {\textbf{Pass\%}} \\
\midrule
Qwen3-4B-it                        & 55 & 59 & 92.73 & 36 & 45 & 86.11 \\
\hspace*{1em}\(\mathbf{R}_{\text{MBPP+} \setminus \text{WikiText-2}}\) & 44 & 56 & 97.73 & 33 & 46 & 81.82 \\
\rowcolor[RGB]{236,244,252}\hspace*{1em}\(\mathbf{R}_{\text{PyComp-1K} \setminus \text{WikiText-2}}\) & \multicolumn{1}{c}{\(16_{\color{red!50!black}\scriptscriptstyle-70.9\%}\)} & \multicolumn{1}{c}{\(18_{\color{red!50!black}\scriptscriptstyle-69.5\%}\)} & 87.50 & \multicolumn{1}{c}{\(14_{\color{red!50!black}\scriptscriptstyle-61.1\%}\)} & \multicolumn{1}{c}{\(19_{\color{red!50!black}\scriptscriptstyle-57.8\%}\)} & 85.71 \\
\bottomrule
\end{tabular}
}
\caption{
Fine-grained suppression of Python comprehension generation on Qwen3-4B-it.
\textbf{Records}: outputs containing \(\geq 1\) comprehension.
\textbf{Total}: aggregate comprehension instances.
\textbf{Pass\%}: pass@1 functional correctness on the subset of generated solutions that use comprehensions.}\label{tab:comprehension_suppression}
\end{table}

To test RACE's resolution on narrow stylistic behaviors, we target Qwen3-4B-it's generation of Python comprehensions.
These are semantically optional but syntactically idiomatic constructs.

Using PyComp-1K as the narrow scoring set, we apply \(\mathbf{R}_{\text{PyComp-1K} \setminus \text{WikiText-2}}\) and \(\mathbf{R}_{\text{MBPP+} \setminus \text{WikiText-2}}\) with a small intervention budget of \(K_{\mathrm{sel}}{=}10\) neurons per layer.

Suppressing PyComp-1K neurons drastically reduces comprehension usage (-70.9\% on MBPP+, -61.1\% on HumanEval+) while largely preserving functional correctness (Table~\ref{tab:comprehension_suppression}).
This decline without catastrophic forgetting suggests that RACE-selected neurons steer behavior rather than merely disrupt it.
The generation-level effect of this targeted perturbation suggests that RACE can use a deliberately designed dataset to localize a specific model behavior.
Appendix~\ref{sec:appendix_pycomp_output_examples} provides a comparison of model outputs before and after perturbation.

\subsubsection{Token-Position Protocol for Observation Collection}\label{subsubsec:token_position_ablation}

\begin{table}[ht]
\centering
\scriptsize
\setlength{\tabcolsep}{4pt}
\resizebox{\columnwidth}{!}{
\begin{tabular}{lccc}
\toprule
\textbf{Benchmark} & \textbf{No suppression} & \textbf{All generated tokens (default)} & \textbf{First-token-only} \\
\midrule
MBPP+\indistmark   & 82.28 & 3.17  & 70.37 \\
HumanEval+\oodmark & 83.54 & 1.22  & 76.83 \\
MATH-500           & 94.40 & 85.41 & 95.00 \\
AMC                & 87.31 & 75.11 & 85.08 \\
MMLU-Redux         & 81.37 & 82.35 & 81.74 \\
GPQA               & 45.45 & 52.53 & 42.42 \\
\bottomrule
\end{tabular}
}
\caption{
Token-position ablation on Qwen3-4B-it under \(\mathbf{R}_{\text{MBPP+} \setminus \text{WikiText-2}}\) with per-layer top-\(1\%\) MLP suppression: accuracy (\%) when observations are collected over all generated-token positions (default) or restricted to the first generated-token position per prompt.}\label{tab:token_position_ablation}
\end{table}

The default observation-collection strategy (Appendix~\ref{sec:appendix_evidence_strategies}) includes every generated-token position in \(T_c\) and records \(e_j^{(t)}\) at each position for each scoring prompt.
We evaluate an alternative token-position choice: a \emph{first-token-only} variant that restricts the observation set to the first generated-token position per prompt, holding all other settings fixed.
Table~\ref{tab:token_position_ablation} compares the resulting benchmark accuracies against the default strategy and the no-suppression baseline.

Restricting the observation set to the first generated-token position substantially weakens the intervention; we therefore posit that the number of observations used for evidence computation critically affects the final auditing accuracy.

\subsection{Computational Efficiency}\label{subsec:computational_efficiency}

\begin{table}[ht]
\centering
\scriptsize
\setlength{\tabcolsep}{4pt}
\begin{tabular}{lrr}
\toprule
\textbf{Method} & \textbf{Extra FLOPs} & \textbf{Forward-equivalent overhead} \\
\midrule
GxAct & \(18.77\) TFLOPs & \(144.29\times\) \\
AttnLRP & \(18.82\) TFLOPs & \(144.62\times\) \\
\rowcolor[RGB]{236,244,252} RDA & \(81.5\) GFLOPs & \(0.63\times\) \\
\bottomrule
\end{tabular}
\caption{Profiler-measured scoring overhead relative to forward inference on Qwen3-4B-it.}\label{tab:computational_cost}
\end{table}

Table~\ref{tab:computational_cost} reports the additional FLOPs required by each scoring method, measured with a profiler on Qwen3-4B-it using \(64\) input tokens and \(16\) generated tokens.
Appendix~\ref{sec:appendix_profiler_scoring_cost} gives the profiling protocol.

\subsection{Discussion}\label{subsec:discussion}

We analyze how scoring-set size and variance regularization affect RACE, using Qwen3-4B-it with MBPP+ unless stated otherwise.

\begin{figure}[t]
\centering
\includegraphics[width=\columnwidth]{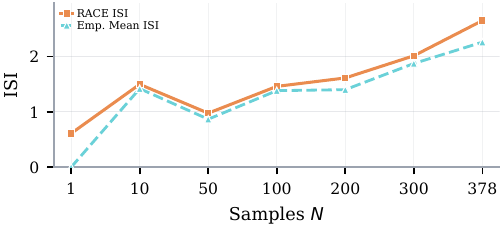}
\caption{
Sample efficiency on Qwen3-4B-it under \(\mathbf{R}_{\text{MBPP+} \setminus \text{WikiText-2}}\): ISI after suppressing top-\(1\%\) MLP neurons per layer as the scoring-set size \(N\) varies, comparing RACE with Emp. Mean.
}\label{fig:hyperparam_sensitivity}
\end{figure}

\noindent\textbf{Scoring-set sample size \(N\).}
The scoring-set sample size \(N\) controls how reliably RACE estimates intervention-worthy neurons from the target domain.
Figure~\ref{fig:hyperparam_sensitivity} evaluates this effect through downstream ISI rather than ranking correlation: for each \(N\), we construct \(\mathbf{R}_{\text{MBPP+} \setminus \text{WikiText-2}}\), suppress the top-\(1\%\) MLP neurons per layer, and compare RACE with Emp. Mean.
The ISI curves are non-monotonic under small and mid-sized scoring sets, but both methods improve as more MBPP+ examples are used and reach their strongest selectivity on the full scoring set (\(N{=}378\)).
Notably, under the default prior, the posterior mean is a monotonic rescaling of the empirical mean (\(\mu_{n,j} \propto \bar{e}_j\)).
The finite-sample difference between RACE and empirical averaging therefore arises from CAM's uncertainty penalty.

\noindent\textbf{Bayesian variance regularization.}
The sample-size behavior clarifies that RACE's low-data advantage comes from calibrated uncertainty rather than a different mean estimator.
Sparse module evidence traces can make empirical variance-based ratios over-rank weak nonzero fluctuations, consistent with the poor MLP intervention specificity of Emp. SNR in Table~\ref{tab:code_suppression}.
CAM avoids this failure mode by retaining a finite uncertainty margin under the default NIG prior.
It becomes prior-insensitive once the induced observation count \(n\) is large.
Appendix~\ref{sec:appendix_variance_regularization} gives the short derivation.

\section{Related Work}\label{sec:related}

Transformer computation is often analyzed as a sequence of additive writes to the residual stream~\citep{elhage2021mathematical, geva2021transformer}.
Learned lenses provide a complementary view by decoding hidden states at intermediate depths~\citep{belrose2023tuned}.
RDA uses the residual decomposition to project each neuron write onto the observation-specific update direction of its host module, defining the score directly in residual space at the current depth.
Prior work has shown that vocabulary-based attribution can be confounded by memory-management writes~\citep{janiak2024adversarial} and downstream compensation after ablation~\citep{mcgrath2023hydra}.

Most behavior-localization methods analyze individual inputs through attribution~\citep{sundararajan2017axiomatic, achtibat2024attnlrp}, causal intervention, or circuit discovery~\citep{kevin2022locating, conmy2023towards}.
Domain-level rankings are typically constructed by averaging these per-input scores.
This aggregation retains the mean contribution while discarding variation across inputs, so similar averages may reflect either broadly distributed contributions or responses concentrated on a few inputs.
Statistical tests for localization have been proposed to assess this distinction and the reliability of the resulting claims~\citep{adebayo2018sanity, shi2024hypothesis}.

Prior work associates neurons with skills~\citep{wang-etal-2022-finding-skill, song-etal-2024-large}, languages~\citep{tang-etal-2024-language}, and factual knowledge~\citep{damai2021knowledge} by thresholding averaged statistics.
Evidence of cross-phenomenon overlap suggests that the resulting sets can mix target-specific and broadly shared responses~\citep{niu2024knowledge}.
The same aggregation issue arises in neuron rankings used for pruning~\citep{sun2024wanda} and steering~\citep{turner2024activation}.
Sparse dictionaries provide finer-grained units~\citep{bricken2023monosemanticity}, although applying them across a model requires separate autoencoders for the audited layers, and recent comparisons report mixed gains over direct probing~\citep{kantamneni2025saeprobing}.
RACE works in the native neuron basis and uses forward-pass evidence to rank each neuron by a conservative lower credible bound on its mean residual alignment.

\section{Conclusion}\label{sec:conclusion}

To assess functional consistency, RACE recasts neuron analysis as inference over latent alignment distributions.
Its posterior score favors neurons whose residual contributions remain positive and stable across observations from a target domain, and suppression experiments show that these neurons affect the corresponding model behaviors.
RACE collects and aggregates evidence in a single streaming pass, so for a fixed model, its scoring cost grows linearly with the
number of token-position observations.
This scaling makes domain-level neuron analysis practical on the real-world models.

\section*{Limitations}
\label{sec:limitations}

We discuss the currently identified limitations in the formulation of RACE.
The extraction of alignment evidence relies on linear residual-stream projections, which means the framework may miss complex non-linear synergies among multiple neurons or distributed polysemantic features that require non-linear decoding.
Additionally, RACE and the other evaluated baselines identify weaker domain-specific effects in attention modules than in MLPs.
One tentative explanation is that attention output channels are more broadly shared across domains: Appendix~\ref{sec:appendix_cross_domain_consistency} reports an all-domain Jaccard overlap of \(0.264\) for the Top-\(1\%\) attention neurons, compared with \(0.085\) for MLP neurons.
Because RSF filters neurons that also score highly on the reference distribution, this greater cross-domain overlap may cause it to remove more broadly reusable attention signal, leaving a weaker domain-specific signal than in MLPs.
Accordingly, the current RSF formulation is less effective at isolating domain-specific attention channels, where task-relevant signals appear to be more entangled with broadly shared functionality.
Developing attention-specific auditing and reference-filtering strategies remains an important direction for future work.

\section*{Ethical Considerations}
\label{sec:ethics}

RACE has dual-use implications: it supports benign model auditing and targeted pruning, but its capability-suppression mechanism could also be used to selectively degrade model capabilities or circumvent safety alignment.
We therefore recommend evaluating model capabilities both before and after suppression and restricting the use of suppression in production systems to authorized auditing.

\ifanonymouspaper
\else
  \section*{Acknowledgements}

  This work was supported in part by the National Natural Science Foundation of China under Grants 52202496, 52442218, and U2433216; The Key Research and Development Project of Nantong City, China (Special Project for Prospective Technology Innovation, No. GZ2024001); and the Key Laboratory of Target Cognition and Application Technology (2023-CXPT-LC-005).
\fi

\bibliography{main}

\clearpage
\appendix
\section{Why Use a Module-Local Axis}
\label{sec:appendix_module_local_axis}

A natural alternative to RDA is to score every neuron against a single global direction derived from the final next-token prediction.
For example, one could replace the module-local axis in Eq.~\eqref{eq:eval_axis} with a normalized unembedding or contrastive logit direction \(\hat{\mathbf{g}}^{(t)}\), and score a neuron by
\begin{equation}
    e_{\mathrm{global},j}^{(t)} = a_j^{(t)} \mathbf{w}_j^\top \hat{\mathbf{g}}^{(t)},
\end{equation}
where \(\mathbf{w}_j\) is the neuron's output vector and \(a_j^{(t)}\) is its activation at observation \(t\).
This resembles vocabulary-projection analyses such as the logit lens and direct logit attribution.
RACE uses the module-local residual update \(\Delta\mathbf{r}_{l,m}^{(t)}\) as the evaluation direction.
The resulting score tracks how consistently a neuron contributes to the update produced at that layer.
A global-logit score tracks the neuron's alignment with the final prediction.

\paragraph{Computations vary across depth.}
Prior work repeatedly shows that Transformer layers are functionally stratified, with different computations emerging at different depths.
In BERT, lower layers capture phrase-level or surface information, middle layers capture syntactic structure, and upper layers capture more semantic information~\citep{jawahar2019what}.
Tenney et al.\ similarly find a localized progression resembling the classical NLP pipeline, from POS tagging and parsing through semantic roles and coreference~\citep{ian2019bert}.
For decoder language models, Geva et al.\ show that FFN memories differ across depth: lower layers tend to match shallower patterns, while upper layers encode more semantic patterns and more directly induce output-vocabulary distributions~\citep{geva2021transformer}.
Follow-up work further views FFN outputs as additive updates to a changing vocabulary distribution~\citep{geva2022transformer}, and factual-recall analyses identify distinct early-MLP enrichment and later information-routing phases~\citep{geva2023dissecting}.
These results imply that an intermediate neuron can be important because it constructs, routes, erases, or transforms information that is not yet aligned with the final answer token.
A final-logit axis therefore imposes a late-stage semantic criterion on layers whose local role may be lexical, syntactic, relational, or preparatory.

\paragraph{Raw logit projections track vocabulary readability across depth.}
Lens-style methods expose how predictions evolve across depth and show why the final unembedding provides an imperfect common coordinate system.
The tuned lens was introduced as a refinement of the logit lens because the raw logit lens is often brittle; the tuned lens learns a separate affine translator for each layer and is reported to produce more predictive, reliable, and less biased intermediate predictions~\citep{belrose2023tuned}.
Patchscopes reaches a similar conclusion from another direction: many vocabulary-projection methods can be viewed as special cases of a broader representation-inspection framework, and their shortcomings include failures in early layers and limited expressivity~\citep{ghandeharioun2024patchscopes}.
The need for layer-specific translators and richer patching contexts indicates that intermediate states require depth-dependent decoding.
Raw alignment with the final unembedding therefore probes when a layer becomes linearly readable in vocabulary space.

\paragraph{Direct logit attribution can mis-rank intermediate components.}
Direct logit attribution projects intermediate residual components onto final logit directions, but this projection ignores the fact that later layers can overwrite, rotate, or erase earlier residual directions.
Janiak et al.\ provide a concrete adversarial example in GELU-4L: the model uses a memory-management mechanism in which later heads and MLPs remove directions written by earlier heads, and DLA becomes misleading because it does not account for this erasure~\citep{janiak2024adversarial}.
For RACE, this failure mode is especially relevant.
A globally positive logit projection at layer \(l\) may be a transient direction that later modules remove, while a globally weak projection may be a necessary intermediate feature that later modules transform into the final prediction.
Ranking neurons by final-logit projection would therefore mix three factors: local functional contribution, survival through subsequent computation, and proximity to the output head.

\paragraph{Global axes conflate alignment magnitude with depth.}
Because later representations are closer to the output distribution, a global logit axis tends to favor late-layer neurons whose effects have already been rotated into vocabulary-readable directions.
This creates a numerical depth bias: the sorted list of ``important'' features can be dominated by late-layer units even when earlier and middle layers contain indispensable local computations.
Such a ranking confounds two quantities that RACE keeps separate: the strength of a neuron's contribution to its module's current update, and the downstream fate of that update after many additional nonlinear transformations.

\paragraph{Implication for RACE.}
The module-local axis provides a layer-specific alignment signal for each observation.
RACE aggregates these signals to estimate population-level functional consistency.
Final behavioral relevance is then tested separately through targeted suppression and reference-set filtering.
This separation matches the evidence from prior work: intermediate layers perform different computations, raw logit projections require careful layer-specific interpretation, and direct logit attribution can be misleading when later layers erase or transform earlier residual directions.

\section{Why Residual Alignment Measures Functional Role}
\label{sec:appendix_residual_alignment_justification}

We first establish two formal properties of the RDA evidence and then explain how they connect per-observation geometry to domain-level functional consistency.

\paragraph{Formal properties of the alignment evidence.}
For a fixed observation, we omit the superscript \((t)\) in the following geometric argument.
Writing each neuron's contribution as \(\mathbf{v}_j=a_j\mathbf{w}_j\) gives the exact decomposition \(\Delta\mathbf{r}=\sum_j\mathbf{v}_j\).
With the realized output direction \(\hat{\mathbf{d}}=\Delta\mathbf{r}/\lVert\Delta\mathbf{r}\rVert_2\) from Eq.~\ref{eq:eval_axis}, the RDA score is \(e_j=\langle\mathbf{v}_j,\hat{\mathbf{d}}\rangle\).
Each neuron write can be decomposed into a component aligned with \(\hat{\mathbf{d}}\) and an orthogonal component:
\begin{equation}
    \begin{aligned}
        \mathbf{v}_j &= e_j\hat{\mathbf{d}}+\mathbf{q}_j,
        &\mathbf{q}_j &\perp\hat{\mathbf{d}},\\
        \sum_j\mathbf{q}_j &= \mathbf{0},
        &\sum_j e_j &= \lVert\Delta\mathbf{r}\rVert_2,
    \end{aligned}
    \label{eq:race_completeness}
\end{equation}
The alignment scores therefore provide an exact partition of the module-output norm.
This identity defines their scope as a module-local geometric quantity.
The same score also gives the local sensitivity of this norm to rescaling neuron \(j\).
For \(\Delta\mathbf{r}_j(\alpha)=\Delta\mathbf{r}+(\alpha-1)\mathbf{v}_j\),
\begin{equation}
    \left.\frac{\partial}{\partial\alpha}\lVert\Delta\mathbf{r}_j(\alpha)\rVert_2\right|_{\alpha=1}=e_j,
    \label{eq:race_sensitivity}
\end{equation}
Thus, \(e_j\) measures the first-order change in the norm of the realized module update when the contribution of neuron \(j\) is rescaled.

\paragraph{Residual writes mediate module communication.}
In the residual architecture considered here, modules pass information to subsequent layers through the residual stream.
A neuron's vector-valued write \(\mathbf{v}_j\) mediates its downstream effect through the residual stream~\citep{elhage2021mathematical,geva2021transformer,geva2022transformer}.
Projecting \(\mathbf{v}_j\) onto \(\hat{\mathbf{d}}\) measures the component of the neuron write that contributes to the realized module update.
The scalar magnitude \(|a_j|\) records activation strength, and the projection supplies the directional information needed to distinguish support, opposition, and orthogonality to that update.

\paragraph{Orthogonal components under feature superposition.}
Under feature superposition, individual neuron writes need not align with the net module update~\citep{elhage2022superposition}.
Their orthogonal components \(\mathbf{q}_j\) may cancel when the writes are summed, as shown in Eq.~\ref{eq:race_completeness}.
RDA retains the projection that contributes to the realized update and excludes the cancelling orthogonal component.
A neuron can therefore have a large activation on a domain input while making little aligned contribution if its output lies primarily in directions cancelled by other neurons.
The sign of \(e_j\) further distinguishes neurons that reinforce the realized update from those that oppose it.

\paragraph{From per-observation alignment to a population-level role.}
A domain-level functional role requires a neuron to contribute consistently across observations.
Strong responses confined to a few inputs provide insufficient evidence for such a role.
The generative model in Eq.~\ref{eq:generative} estimates this population-level property from the observed alignment evidence.
A neuron that repeatedly contributes in the positive module-update direction produces positive \(e_j^{(t)}\) values with limited dispersion.
CAM combines the posterior mean \(\mu_{n,j}\) with the uncertainty determined in part by \(\mathrm{SS}_j\), as defined in Eq.~\ref{eq:cam}.
Magnitude therefore corresponds to the average signed contribution to the realized module update, while stability corresponds to the reproducibility of that contribution across observations.
Evidence that appears on only a few inputs or changes sign across inputs increases dispersion and reduces the CAM score even when the neuron has a large peak activation.
RDA supplies the per-observation evidence, and Bayesian aggregation converts repeated positive alignment into a domain-level consistency score.

\paragraph{The module increment as the evaluation axis.}
The identities in Eqs.~\ref{eq:race_completeness}--\ref{eq:race_sensitivity} depend on using the module increment \(\Delta\mathbf{r}\) as the evaluation axis.
The full residual \(\mathbf{r}_{\mathrm{in}}+\Delta\mathbf{r}\) contains the accumulated outputs of previous layers, so projecting onto it would mix the current module's contribution with computations performed upstream.
Using the module increment preserves the identity \(\sum_j e_j=\lVert\Delta\mathbf{r}\rVert_2\) and restricts the alignment measure to the update produced by the current module.
Targeted suppression and reference-set filtering evaluate how these local alignment signals affect final model outputs.

\section{RACE Observation-Collection Protocols}
\label{sec:appendix_evidence_strategies}

RACE supports two primary observation-collection protocols based on the nature of the evaluation corpus:
\begin{itemize}
    \item \textbf{Autoregressive Observations:} For task-solving benchmarks like MBPP+ and MATH-500, we perform full autoregressive generation.
    Generated-token positions are included in \(T_c\), and RDA records \(e_j^{(t)}\) at each position.
    The prompt/prefill positions provide necessary conditioning but are not included in the target-domain observation set, focusing the scoring on the model's active generation behavior.
    \item \textbf{Teacher-Forced Observations:} For continuous text corpora (WikiText-2) or fixed structural behaviors (PyComp-1K), we execute a single forward pass over the provided text sequences using teacher forcing.
    In this setting, all token positions within the sequence are included in \(T_c\), and RDA records \(e_j^{(t)}\) at each position.
    This strategy relies on the intrinsic sequence structure without requiring the model to sequentially produce new tokens.
\end{itemize}

Under the autoregressive protocol, the Qwen3-4B-it auditing runs induce token-position observation counts of \(n=705{,}426\) for MATH-500 and \(n=98{,}560\) for MBPP+.

\section{Cross-Domain Consistency of RACE-Selected Neurons}
\label{sec:appendix_cross_domain_consistency}

This appendix analyzes whether neurons selected by RACE at a fixed top fraction \(\tau_{\mathrm{sel}}\) are shared across distinct task domains.
The goal is to distinguish broadly shared selected neurons from neurons selected primarily for a single target distribution.
We evaluate three domains: mathematical reasoning, code generation, and general language modeling, as summarized in Table~\ref{tab:cross_domain_datasets}.

\begin{table}[ht]
\centering
\scriptsize
\setlength{\tabcolsep}{5pt}
\begin{tabular}{lll}
\toprule
\textbf{Domain} & \textbf{Type} & \textbf{Description} \\
\midrule
\texttt{Math-500} & Reasoning & Mathematical problem solving \\
\texttt{MBPP+} & Code & Python programming tasks \\
\texttt{WikiText-2} & Language & General text modeling \\
\bottomrule
\end{tabular}
\caption{Domains used for cross-domain consistency analysis.}
\label{tab:cross_domain_datasets}
\end{table}

\paragraph{Analyzed Modules.}
For each configuration, we analyze both ATTN and MLP.
For each layer and module, RACE ranks neurons by \texttt{CAM}.
We then select the top \(\tau_{\mathrm{sel}}\) fraction of neurons, where \(\tau_{\mathrm{sel}} \in \{1\%, 5\%, 10\%\}\).

\paragraph{Metrics.}
Let \(S_d^{(l,m,\tau_{\mathrm{sel}})}\) denote the neuron set selected at fraction \(\tau_{\mathrm{sel}}\) for domain \(d\), layer \(l\), and module \(m\).
For each pair of domains \(d_1,d_2\), we compute the layer-wise Jaccard similarity:
\begin{equation}
    J(d_1,d_2) =
    \frac{|S_{d_1}^{(l,m,\tau_{\mathrm{sel}})} \cap S_{d_2}^{(l,m,\tau_{\mathrm{sel}})}|}
         {|S_{d_1}^{(l,m,\tau_{\mathrm{sel}})} \cup S_{d_2}^{(l,m,\tau_{\mathrm{sel}})}|}.
\end{equation}
We additionally compute an all-domain overlap score that requires a neuron to be shared by all three domains:
\begin{equation}
    J_{\mathrm{all}} =
    \frac{|S_{\mathrm{math500}} \cap S_{\mathrm{mbpp\_plus}} \cap S_{\mathrm{wikitext2}}|}
         {|S_{\mathrm{math500}} \cup S_{\mathrm{mbpp\_plus}} \cup S_{\mathrm{wikitext2}}|}.
\end{equation}
All reported values are averaged over layers, with standard deviations computed across layers.

\paragraph{Main Finding.}
Attention neurons are substantially more domain-general than MLP neurons.
Under \texttt{CAM} with a Top-\(1\%\) threshold, attention neurons obtain an all-domain Jaccard of \(0.264\) (\(\mathrm{std}=0.099\)), meaning that roughly \(26\%\) of the selected attention neurons are shared across all three domains.
In contrast, MLP neurons obtain an all-domain Jaccard of only \(0.085\) (\(\mathrm{std}=0.052\)), meaning that only about \(8.5\%\) of the selected MLP neurons are shared.
This gives a \(3.1\times\) gap between attention and MLP modules.

\begin{table}[ht]
\centering
\scriptsize
\setlength{\tabcolsep}{4pt}
\begin{tabular}{lccc}
\toprule
\textbf{Configuration} & \textbf{Attention} & \textbf{MLP} & \textbf{Ratio} \\
\midrule
\texttt{CAM}, Top-\(1\%\) & \(0.264\) & \(0.085\) & \(3.1\times\) \\
\texttt{CAM}, Top-\(5\%\) & \(0.249\) & \(0.110\) & \(2.3\times\) \\
\texttt{CAM}, Top-\(10\%\) & \(0.258\) & \(0.137\) & \(1.9\times\) \\

\bottomrule
\end{tabular}
\caption{
All-domain Jaccard similarity of neuron sets selected at the same fraction \(\tau_{\mathrm{sel}}\) under \(\mathbf{R}_{\text{MATH-500}}\), \(\mathbf{R}_{\text{MBPP+}}\), and \(\mathbf{R}_{\text{WikiText-2}}\).
Attention neurons are consistently more shared across domains than MLP neurons.}
\label{tab:cross_domain_all_jaccard}
\end{table}

Table~\ref{tab:cross_domain_all_jaccard} shows that the gap is robust across both scoring metrics and all selection fractions \(\tau_{\mathrm{sel}}\).
As \(\tau_{\mathrm{sel}}\) increases, the MLP all-domain overlap rises moderately, but attention remains consistently higher.
These results suggest that attention output channels contain a larger population of reusable cross-domain routing or integration neurons.
MLP down-projection neurons show lower cross-domain overlap and narrower response patterns under RACE scoring.

\begin{figure*}[t]
\centering
\includegraphics[width=0.85\textwidth]{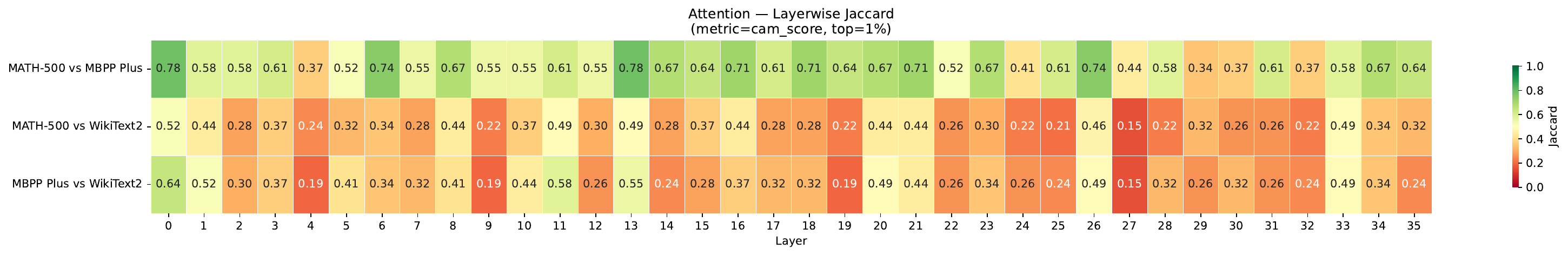}

\includegraphics[width=0.85\textwidth]{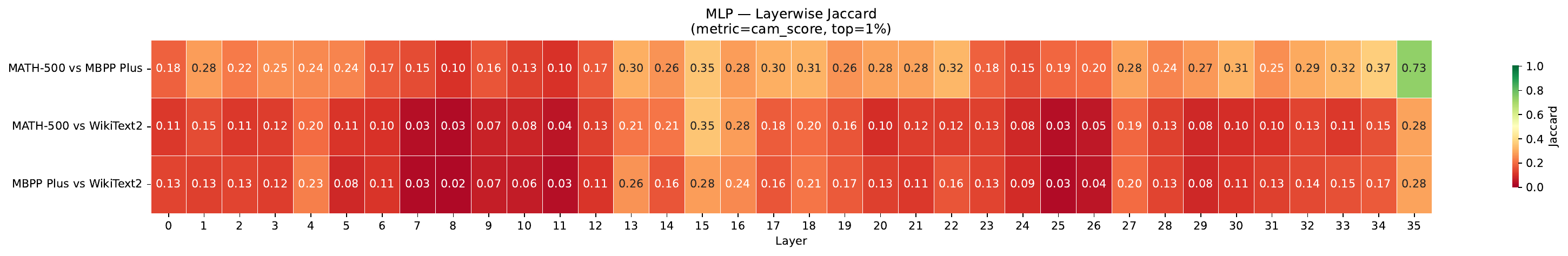}
\caption{
\textbf{Layer-wise cross-domain overlap at selection fraction \(\tau_{\mathrm{sel}}{=}1\%\) (CAM score).}
Heatmaps show Jaccard similarity between domain pairs across transformer layers for the selected module, attribution metric, and selection fraction.
Each cell reports the overlap between the two domains' top-ranked neuron sets in that layer, with higher values indicating greater cross-domain consistency in the identified important neurons.
\textbf{Top:} attention neurons exhibit substantially higher cross-domain overlap, consistent with the all-domain Jaccard of \(0.264\).
\textbf{Bottom:} MLP neurons show markedly lower inter-domain agreement, reflecting more domain-specific response patterns (all-domain Jaccard \(0.085\)).}\label{fig:cross_domain_heatmap}
\end{figure*}

\section{Math Domain Intervention with Reference-Set Filtering on Llama-3.1-8B-it}
\label{sec:appendix_math_reference_filtered}

This appendix reports additional \(K_{\mathrm{sel}}{=}50\) ATTN and MLP suppression results for \(\mathbf{R}_{\text{MATH-500} \setminus \text{WikiText-2}}\) on Llama-3.1-8B-it in Table~\ref{tab:math_suppression_reference_filtered}.
Neurons are scored on MATH-500, and the intervention set removes neurons that overlap with a reference RACE result computed on WikiText-2 before suppression.

\begin{table}[ht]
\centering
\scriptsize
\setlength{\tabcolsep}{3pt}
\resizebox{\linewidth}{!}{
\begin{tabular}{c@{\hspace{3pt}}l@{\hspace{3pt}}S[table-format=2.2]@{\hspace{5pt}}S[table-format=2.2]@{\hspace{5pt}}S[table-format=2.2]@{\hspace{5pt}}S[table-format=2.2]@{\hspace{5pt}}S[table-format=2.2]@{\hspace{5pt}}S[table-format=1.2]}
\toprule
\textbf{Module} & \textbf{Method} & {\textbf{MATH-500\indistmark} \(\downarrow\)} & {\textbf{AMC\oodmark} \(\downarrow\)} & {\textbf{GPQA}} & {\textbf{MMLU-Redux}} & {\textbf{ARC}} & {\textbf{ISI}\(\uparrow\)} \\
\midrule
\multirow{2}{*}{ATTN}
 & Neg. CAM & 47.00 & 15.67 & 28.79 & 71.35 & 86.10 & 2.02 \\
 & \textbf{RACE}   & 27.00 & 6.72 & 22.73 & 67.35 & 86.16 & 1.65 \\
\midrule
\multirow{2}{*}{MLP}
 & Neg. CAM & 51.00 & 23.14 & 31.31 & 72.61 & 86.22 & 0.98 \\
 & \textbf{RACE}   & 8.80 & 2.24 & 24.75 & 69.89 & 85.71 & 2.36 \\
\midrule
\multicolumn{2}{l}{Llama-3.1-8B-it} & 50.40 & 24.63 & 29.80 & 72.91 & 86.16 & {---} \\
\bottomrule
\end{tabular}
}
\caption{Math domain with \(\mathbf{R}_{\text{MATH-500} \setminus \text{WikiText-2}}\): accuracy on Llama-3.1-8B-it after suppressing the top \(K_{\mathrm{sel}}{=}50\) neurons per layer.}\label{tab:math_suppression_reference_filtered}
\end{table}

\section{Model, Generation, and Evaluation Details}
\label{sec:appendix_eval_details}

This appendix summarizes the model configurations, decoding parameters, and benchmark evaluation protocol used in the experiments.
Benchmark evaluation is conducted with \texttt{EvalScope}~1.5.0 using \texttt{vLLM}~0.15.1 as the inference backend.
Unless otherwise specified, benchmark scores follow the official metric implementation exposed by \texttt{EvalScope}.
We report the resulting benchmark score as Domain Accuracy (DA) in the main tables.
Model generation settings, architectural dimensions, benchmark roles, EvalScope benchmark metadata, and corpus statistics are reported in Tables~\ref{tab:appendix_model_generation}, \ref{tab:appendix_model_architecture}, \ref{tab:appendix_eval_setup}, \ref{tab:appendix_evalscope_benchmark_metadata}, \ref{tab:appendix_evalscope_benchmark_statistics}, and~\ref{tab:appendix_non_evalscope_corpus_statistics}.

\begin{table}[ht]
\centering
\scriptsize
\setlength{\tabcolsep}{4pt}
\resizebox{\linewidth}{!}{
\begin{tabular}{lrrrr}
\toprule
\textbf{Model} & \textbf{Scale} & \textbf{Temperature} & \textbf{Top-\(p_{\mathrm{dec}}\)} & \textbf{Top-\(K_{\mathrm{dec}}\)} \\
\midrule
Qwen3-4B-it-2507 & 4B & 0.7 & 0.8 & 20 \\
Llama-3.1-8B-it & 8B & 0.6 & 0.9 & {---} \\
OLMo-3.1-32B-it & 32B & 0.6 & 0.95 & {---} \\
\bottomrule
\end{tabular}
}
\caption{
Model and decoding settings used for benchmark evaluation.
Here \(p_{\mathrm{dec}}\) and \(K_{\mathrm{dec}}\) denote the nucleus-sampling probability and integer top-\(K\) decoding cutoff, respectively.
Parameters not listed are left at the evaluator or backend default; ``---'' indicates that the parameter is unset.}
\label{tab:appendix_model_generation}
\end{table}

\begin{table}[ht]
\centering
\scriptsize
\setlength{\tabcolsep}{4pt}
\resizebox{\linewidth}{!}{
\begin{tabular}{lrrr}
\toprule
\textbf{Model} & \textbf{Layers} & \textbf{ATTN neurons / layer} & \textbf{MLP neurons / layer} \\
\midrule
Qwen3-4B-it-2507 & 36 & 2560 & 9728 \\
Llama-3.1-8B-it & 32 & 4096 & 14336 \\
OLMo-3.1-32B-it & 64 & 5120 & 27648 \\
\bottomrule
\end{tabular}
}
\caption{
Architectural dimensions of all evaluated models.}
\label{tab:appendix_model_architecture}
\end{table}

\begin{table}[ht]
\centering
\scriptsize
\setlength{\tabcolsep}{5pt}
\resizebox{\linewidth}{!}{
\begin{tabular}{llll}
\toprule
\textbf{Benchmark} & \textbf{Role} & \textbf{Reported score} & \textbf{Evaluator} \\
\midrule
MBPP+ & Code scoring / in-distribution validation & EvalScope benchmark score & EvalScope + vLLM \\
HumanEval+ & Code out-of-distribution validation & EvalScope benchmark score & EvalScope + vLLM \\
MATH-500 & Math scoring / in-distribution validation & EvalScope benchmark score & EvalScope + vLLM \\
AMC & Math out-of-distribution validation & EvalScope benchmark score & EvalScope + vLLM \\
ARC & Non-target retention / reasoning control & EvalScope benchmark score & EvalScope + vLLM \\
MMLU-Redux & General-knowledge retention & EvalScope benchmark score & EvalScope + vLLM \\
GPQA & General-knowledge retention & EvalScope benchmark score & EvalScope + vLLM \\
\bottomrule
\end{tabular}
}
\caption{
Benchmark evaluation setup.}
\label{tab:appendix_eval_setup}
\end{table}

\begin{table}[ht]
\centering
\scriptsize
\setlength{\tabcolsep}{3pt}
\resizebox{\linewidth}{!}{
\begin{tabular}{lllll}
\toprule
\textbf{Benchmark} & \textbf{EvalScope key} & \textbf{Task type} & \textbf{Split} & \textbf{Metric / aggregation} \\
\midrule
MBPP+ & \texttt{mbpp\_plus} & Python code generation & \texttt{test} & \texttt{acc}; \texttt{mean\_and\_pass\_at\_k} \\
HumanEval+ & \texttt{humaneval\_plus} & Python code generation & \texttt{test} & \texttt{acc}; \texttt{mean\_and\_pass\_at\_k} \\
MATH-500 & \texttt{math\_500} & Mathematical problem solving & \texttt{test} & \texttt{acc} \\
AMC & \texttt{amc} & Competition math multiple choice & N/A & \texttt{acc} \\
ARC & \texttt{arc} & Multiple-choice science QA & \texttt{test} & \texttt{acc} \\
MMLU-Redux & \texttt{mmlu\_redux} & Multiple-choice knowledge QA & \texttt{test} & \texttt{acc} \\
GPQA & \texttt{gpqa\_diamond} & Expert-level science MCQ & \texttt{train} & \texttt{acc} \\
\bottomrule
\end{tabular}
}
\caption{EvalScope benchmark metadata for all benchmark datasets used in the paper, extracted from the corresponding EvalScope benchmark cards~\citep{evalscope_mbppplus,evalscope_humanevalplus,evalscope_math500,amc2023,evalscope_arc,evalscope_mmlu_redux,evalscope_gpqa_diamond}.}
\label{tab:appendix_evalscope_benchmark_metadata}
\end{table}

\begin{table}[ht]
\centering
\scriptsize
\setlength{\tabcolsep}{3pt}
\resizebox{\linewidth}{!}{
\begin{tabular}{lrrrl}
\toprule
\textbf{Benchmark} & \textbf{Samples} & \textbf{Mean prompt length} & \textbf{Min / max prompt length} & \textbf{Subset / coverage notes} \\
\midrule
MBPP+ & 378 & 375.53 & 222 / 2801 & Python programming tasks with expanded tests \\
HumanEval+ & 164 & 609.57 & 274 / 1519 & Original HumanEval problems with enhanced tests \\
MATH-500 & 500 & 266.89 & 91 / 1804 & Levels 1--5: 43 / 90 / 105 / 128 / 134 examples \\
AMC & 134 & 324.58 & 98 / 1218 & AMC22 / AMC23 / AMC24: 43 / 46 / 45 examples \\
ARC & 3548 & 424.43 & 253 / 1157 & ARC-Easy / ARC-Challenge: 2376 / 1172 examples \\
MMLU-Redux & 5700 & 600.81 & 255 / 5082 & 57 subjects; 100 examples per subject \\
GPQA & 198 & 841.15 & 340 / 5845 & GPQA-Diamond expert science subset \\
\bottomrule
\end{tabular}
}
\caption{
Dataset statistics for all EvalScope benchmarks used in the paper.
Prompt lengths are measured in characters as reported by the EvalScope benchmark cards~\citep{evalscope_mbppplus,evalscope_humanevalplus,evalscope_math500,amc2023,evalscope_arc,evalscope_mmlu_redux,evalscope_gpqa_diamond}.}
\label{tab:appendix_evalscope_benchmark_statistics}
\end{table}

\begin{table}[ht]
\centering
\scriptsize
\setlength{\tabcolsep}{4pt}
\resizebox{\linewidth}{!}{
\begin{tabular}{llll}
\toprule
\textbf{Dataset / corpus} & \textbf{Role in paper} & \textbf{Size / statistics} & \textbf{Construction or source notes} \\
\midrule
WikiText-2~\citep{merity2016pointer} & General-language RSF reference & 2,088,628 train tokens; 217,646 validation tokens; 245,569 test tokens & Wikipedia Good/Featured articles; vocabulary of 33,278 words \\
The Stack~\citep{kocetkov2022stack,bigcode_the_stack_docs} & Source corpus for PyComp-1K construction & 6.4 TB permissively licensed source code across 358 programming languages & Only Python files are streamed for PyComp-1K extraction before AST filtering \\
PyComp-1K & Narrow code-behavior scoring corpus & 1000 AST-verified Python statements & Extracted from \texttt{bigcode/the-stack}~\citep{kocetkov2022stack,bigcode_the_stack_docs}; targets \texttt{ListComp}, \texttt{SetComp}, \texttt{DictComp}, and \texttt{GeneratorExp} nodes \\
\bottomrule
\end{tabular}
}
\caption{
Non-EvalScope corpora and locally constructed datasets appearing in the paper.
WikiText-2 provides \(D_{\mathrm{ref}}\) for \(\mathbf{R}_{D_{\mathrm{tar}} \setminus \text{WikiText-2}}\), The Stack~\citep{kocetkov2022stack,bigcode_the_stack_docs} is the source corpus for constructing PyComp-1K, and PyComp-1K provides \(D_{\mathrm{tar}}\) in the fine-grained \(\mathbf{R}_{\text{PyComp-1K} \setminus \text{WikiText-2}}\) intervention.}
\label{tab:appendix_non_evalscope_corpus_statistics}
\end{table}

\section{Computational Architecture}
\label{sec:appendix_computational_architecture}

This appendix reports the compute environments used for the main experiments.
Experiments on Qwen3-4B-it were executed on the 8-GPU NVIDIA RTX 4090 machine.
Experiments on Llama-3.1-8B-it and OLMo-3.1-32B-it were executed on the 8-GPU NVIDIA A100 80GB machine.
The full hardware and system configuration is given in Table~\ref{tab:appendix_compute_architecture}.

\begin{table}[ht]
\centering
\scriptsize
\setlength{\tabcolsep}{3pt}
\resizebox{\linewidth}{!}{
\begin{tabular}{lllll}
\toprule
\textbf{Environment} & \textbf{System} & \textbf{CPU} & \textbf{Memory} & \textbf{GPU / experiments} \\
\midrule
RTX 4090 server & Ubuntu, Linux 6.8.0-107-generic & Intel Xeon Gold 6138 & 125GiB RAM & 8\(\times\) NVIDIA GeForce RTX 4090; Qwen3-4B-it \\
A100 server & Linux 4.18.0-147 & AMD EPYC 7713 & 368GiB RAM & 8\(\times\) NVIDIA A100 80GB; Llama-3.1-8B-it and OLMo-3.1-32B-it \\
\bottomrule
\end{tabular}
}
\caption{
Computational architecture used in the experiments.
CPU, memory, operating system, and GPU information are reported from the execution environments.}
\label{tab:appendix_compute_architecture}
\end{table}

\section{Profiler-Based Scoring Cost Measurement}
\label{sec:appendix_profiler_scoring_cost}

We measure the computational-efficiency numbers in Table~\ref{tab:computational_cost} with \texttt{torch.profiler(with\_flops=True)} on Qwen3-4B-it.
The forward reference is the profiled cost of the same \(64\)-input/\(16\)-output window, including the prefill-to-generation increment and the selected \texttt{lm\_head} projections.
This reference costs \(130.100\) GFLOPs.

\paragraph{Protocol.}
All methods use the same WikiText-2 token window, target positions, target token ids, and target modules.
We use one lookahead token to define the next-token target for the 16th generated position, but do not increase the model input length.
For GxAct and AttnLRP, we batch all target layers together by default, avoiding artificial repetition of the same forward/backward computation once per layer.
The profiler records only the extra scoring computation; CAM, NIG updates, HDF5 writes, and CPU aggregation are excluded.
We use eager attention so that PyTorch's FLOP profiler can observe the attention matrix multiplications instead of hiding them inside fused kernels.

\begin{table}[ht]
\centering
\scriptsize
\setlength{\tabcolsep}{5pt}
\begin{tabular}{lrr}
\toprule
\textbf{Measurement} & \textbf{Profiler FLOPs} & \textbf{Relative to forward} \\
\midrule
Forward reference & \(130.100\) GFLOPs & \(100.000\%\) \\
\textbf{RDA extra} & \(\mathbf{81.545}\) GFLOPs & \(\mathbf{62.679\%}\) \\
GxAct extra & \(18.772\) TFLOPs & \(14428.812\%\) \\
AttnLRP extra & \(18.815\) TFLOPs & \(14461.987\%\) \\
\bottomrule
\end{tabular}
\caption{Profiler-measured FLOPs for Qwen3-4B-it under the \(64\)-input/\(16\)-output protocol.}
\label{tab:appendix_profiler_scoring_cost}
\end{table}

\section{PyComp-1K Dataset}
\label{sec:appendix_py_comp_dataset}

We use PyComp-1K to evaluate model behavior involving specific Python syntactic patterns.
This dataset contains 1,000 Python statements extracted from the Python subset of \texttt{bigcode/the-stack}~\citep{kocetkov2022stack,bigcode_the_stack_docs}.
Each example consists of the nearest enclosing Python statement around one or more comprehension expressions.

\subsection{Construction Process}

The dataset construction process streamed Python files from The Stack~\citep{kocetkov2022stack,bigcode_the_stack_docs} and parsed each source file using Python's \texttt{ast} module.
The extraction logic specifically targeted four types of AST comprehension nodes: \texttt{ListComp}, \texttt{SetComp}, \texttt{DictComp}, and \texttt{GeneratorExp}.
For each unique comprehension span identified, the nearest enclosing statement was extracted and saved.
Rows that could not be parsed as valid Python source were skipped.

\subsection{Dataset Statistics}

The 1,000 extracted statements exhibit the following distribution of comprehension types (note that a single statement may contain multiple comprehensions):

\begin{table}[ht]
\centering
\small
\begin{tabular}{lr}
\toprule
\textbf{Comprehension Type} & \textbf{Count} \\
\midrule
List Comprehension (\texttt{ListComp}) & 754 \\
Generator Expression (\texttt{GeneratorExp}) & 181 \\
Dictionary Comprehension (\texttt{DictComp}) & 76 \\
Set Comprehension (\texttt{SetComp}) & 9 \\
\bottomrule
\end{tabular}
\caption{Distribution of comprehension types in the Python Comprehension Statements 1K dataset.}
\label{tab:appendix_py_comp_stats}
\end{table}

Each row in the dataset provides the extracted statement, the corresponding line numbers in the original source, the set of AST comprehension types present, and the source metadata from The Stack~\citep{kocetkov2022stack,bigcode_the_stack_docs} (e.g., commit hash, repository name, and file path).

\section{Qualitative Outputs under the PyComp-1K Perturbation}
\label{sec:appendix_pycomp_output_examples}

This section reports five representative Python-generation samples under the PyComp-1K perturbation setting.
For each pair, the first block shows a correct unmodified model output, while the second block shows an incorrect output after perturbing RACE-selected PyComp-1K neurons.

\appendixexample{Example 1: square individual list elements.}
\noindent\begin{minipage}[t]{\linewidth}
\textbf{Original model (correct).}
\begin{lstlisting}[style=appendixpython]
```python
def square_nums(nums):
    """
    Takes a list of numbers and returns a new list containing the square of each element.

    Args:
        nums (list): A list of integers or floats

    Returns:
        list: A list where each element is the square of the corresponding element in the input list

    Examples:
        >>> square_nums([1, 2, 3, 4, 5, 6, 7, 8, 9, 10])
        [1, 4, 9, 16, 25, 36, 49, 64, 81, 100]
        >>> square_nums([10, 20, 30])
        [100, 400, 900]
        >>> square_nums([12, 15])
        [144, 225]
    """
    return [num ** 2 for num in nums]
```
\end{lstlisting}
\end{minipage}\par\smallskip
\noindent\begin{minipage}[t]{\linewidth}
\textbf{Perturbed model (incorrect).}
\begin{lstlisting}[style=appendixpython]
```python
def square_nums(nums):
    """
    Takes a list of numbers and returns a new list with the square of each element.

    Args:
        nums (list): A list of integers or floats

    Returns:
        list: A list containing the square of each element from the input list
    """
    return [num**2]
```
\end{lstlisting}
\end{minipage}

\appendixexample{Example 2: extract strings of a specified length.}
\noindent\begin{minipage}[t]{\linewidth}
\textbf{Original model (correct).}
\begin{lstlisting}[style=appendixpython]
```python
def extract_string(string_list, size):
    """
    Extract strings from a list that have a specified length.

    Args:
        string_list: List of strings to filter
        size: The length to check for (strings of this exact length will be returned)

    Returns:
        List of strings that have the specified length
    """
    return [s for s in string_list if len(s) == size]
```
\end{lstlisting}
\end{minipage}\par\smallskip
\noindent\begin{minipage}[t]{\linewidth}
\textbf{Perturbed model (incorrect).}
\begin{lstlisting}[style=appendixpython]
```python
def extract_string(string_list, size):
    """
    Extract strings from a list that have the specified length.

    Args:
        string_list: List of strings to filter
        size: The length to filter by (only strings of this exact length are returned)

    Returns:
        List of strings that have the exact specified length
    """
    return [string for len(string) == size]
```
\end{lstlisting}
\end{minipage}

\appendixexample{Example 3: sort a dictionary by value.}
\noindent\begin{minipage}[t]{\linewidth}
\textbf{Original model (correct).}
\begin{lstlisting}[style=appendixpython]
```python
def sort_counter(d):
    """
    Sort a dictionary by value in descending order and return a list of tuples.

    Args:
        d (dict): A dictionary with keys and values to be sorted

    Returns:
        list: A list of tuples (key, value) sorted by value in descending order
    """
    return sorted(d.items(), key=lambda item: item[1], reverse=True)
```
\end{lstlisting}
\end{minipage}\par\smallskip
\noindent\begin{minipage}[t]{\linewidth}
\textbf{Perturbed model (incorrect).}
\begin{lstlisting}[style=appendixpython]
```python
def sort_counter(dict_input):
    """
    Sort a dictionary of key-value pairs by value in descending order.

    Args:
        dict_input (dict): A dictionary where each key-value pair is a tuple or key-value pair

    Returns:
        list: A list of tuples sorted by value in descending order
    """
    # Convert the dictionary to a list of tuples and sort by value (second element) in descending order
    return sorted(dict_input, key=lambda x: x[1], reverse=True)
```
\end{lstlisting}
\end{minipage}

\appendixexample{Example 4: sorted union of two tuples.}
\noindent\begin{minipage}[t]{\linewidth}
\textbf{Original model (correct).}
\begin{lstlisting}[style=appendixpython]
```python
def union_elements(tuple1, tuple2):
    """
    Find the union of elements from two tuples and return them in sorted order.

    Args:
        tuple1: First tuple of elements
        tuple2: Second tuple of elements

    Returns:
        A tuple containing the union of elements from both tuples, sorted in ascending order
    """
    # Convert tuples to sets to find union (automatically removes duplicates)
    union_set = set(tuple1) | set(tuple2)

    # Convert back to sorted tuple
    sorted_union = tuple(sorted(union_set))

    return sorted_union
```
\end{lstlisting}
\end{minipage}\par\smallskip
\noindent\begin{minipage}[t]{\linewidth}
\textbf{Perturbed model (incorrect).}
\begin{lstlisting}[style=appendixpython]
```python
def union_elements(tuple1, tuple2):
    """
    Find the union of elements from two tuples and return them in sorted order.

    Args:
        tuple1: First tuple of elements
        tuple2: Second tuple of elements

    Returns:
        A tuple containing the union of elements from both input tuples, sorted in ascending order
    """
    # Convert tuples to sets to find union (remove duplicates)
    set1 = set(tuple1)
    set2 = set(tuple2

    # Find union of both sets
    union_set = set1.union(set2)

    # Convert back to sorted tuple
    sorted_union = tuple(sorted(union_set))

    return sorted_union
```
\end{lstlisting}
\end{minipage}

\appendixexample{Example 5: retrieve the \(k\)-th array element.}
\noindent\begin{minipage}[t]{\linewidth}
\textbf{Original model (correct).}
\begin{lstlisting}[style=appendixpython]
```python
def kth_element(arr, k):
    """
    Find the kth element in the given array using 1-based indexing.

    Args:
        arr: List of integers
        k: 1-based index of the element to find

    Returns:
        The kth element in the array

    Raises:
        IndexError: If k is out of bounds (k > length of array)
    """
    if k < 1:
        raise IndexError("k must be a positive integer")

    if k > len(arr):
        raise IndexError(f"k ({k}) is greater than array length ({len(arr)})")

    return arr[k - 1]  # Convert 1-based index to 0-based index
```
\end{lstlisting}
\end{minipage}\par\smallskip
\noindent\begin{minipage}[t]{\linewidth}
\textbf{Perturbed model (incorrect).}
\begin{lstlisting}[style=appendixpython]
```python
def kth_element(arr, k):
    """
    Find the kth element in the given array using 1-based indexing.

    Args:
        arr: List of numbers
        k: 1-based index of the element to find

    Returns:
        The kth element in the array

    Raises:
        IndexError: If k is out of bounds (k < 1 or k > len(arr))
    """
    # Validate input
    if k < 1:
        raise IndexError("k must be a positive integer (1-based indexing)")

    if k > len(arr):
        raise IndexError(f"k ({k}) is greater than array length ({len(arr})")

    # Return the kth element (k-1 index in 0-based indexing)
    return arr[k - 1]
```
\end{lstlisting}
\end{minipage}

\section{Robustness to Prior Settings and Confidence Levels}
\label{sec:appendix_prior_settings}

When the sample size is sufficiently large (\(n \gg \lambda_0\)), the posterior distribution is dominated by empirical statistics, rendering the choice of prior parameters largely inconsequential.
Empirically, across a wide range of prior configurations (\(\lambda_0, \beta_0 \in [10^{-3}, 10^3]\), \(\alpha_0 \in [0.5, 50]\)), the top-\(1\%\) selected neurons remain highly stable (Jaccard similarity \(\ge 0.98\)).
The CAM confidence level \(\gamma\) primarily sets an absolute verification threshold and has little effect on the relative ranking of the top candidates.
Consequently, the neuron sets selected under fixed proportional budgets remain largely identical across different \(\gamma\) values.
In practice, adopting a more stringent confidence level (e.g., tightening \(\gamma\) from \(0.05\) to \(0.001\)) effectively filters out marginal candidates, shrinking the pool of valid positive-CAM neurons from \(81.75\%\) to \(62.24\%\) (at \(N=10\)), without displacing the most prominent neurons.

\section{Distributional Disruption Metrics}
\label{sec:appendix_distributional_metrics}

The main paper evaluates token-distribution-level disruption using relative perplexity degradation \(\Delta_\mathrm{PPL}\) and mean forward Kullback--Leibler divergence \(\bar{D}_{\mathrm{KL}}\).
\(\Delta_\mathrm{PPL}\) is reported in percentage form:
\begin{equation}
  \Delta_\mathrm{PPL}
  = \frac{\mathrm{PPL}_{\mathrm{sup}} - \mathrm{PPL}_{\mathrm{base}}}
         {\mathrm{PPL}_{\mathrm{base}}}.
  \label{eq:delta_ppl}
\end{equation}
This aggregates the increase in per-token surprisal and quantifies the overall deterioration in predictive quality.
\(\bar{D}_{\mathrm{KL}}\) is computed as
\begin{equation}
  \begin{aligned}
  \bar{D}_{\mathrm{KL}}
  &= \frac{1}{T}\sum_{p=1}^{T}
  D_{\mathrm{KL}}\!\bigl(
    P_{\text{base}}(\cdot\mid\mathbf{x}_{<p}) \,\| \\
  &\hspace{4em}
    P_{\text{sup}}(\cdot\mid\mathbf{x}_{<p})
  \bigr).
  \end{aligned}
  \label{eq:mean_kl}
\end{equation}
This measures the mean distributional shift per token position and can detect behavioral changes even when aggregate likelihood is largely preserved.

\section{Without RSF (w/o-RSF) Distributional Disruption on Qwen3-4B-it}
\label{sec:appendix_no_rsf_distributional_qwen3}

This appendix reports distributional disruption results w/o-RSF for Qwen3-4B-it.
All checkpoints suppress the top \(K_{\mathrm{sel}}{=}5\) RACE-selected neurons per layer.
They are evaluated by direct model comparison against the unmodified model on the first 100 samples per dataset in order.
The resulting \(\Delta_\mathrm{PPL}\) and \(\bar{D}_{\mathrm{KL}}\) measurements are summarized in Table~\ref{tab:appendix_no_rsf_ppl_kl_qwen3}.

\begin{table}[ht]
\centering
\scriptsize
\setlength{\tabcolsep}{3pt}
\resizebox{\linewidth}{!}{
\begin{tabular}{ll@{\hspace{6pt}} c c c @{\hspace{12pt}} c c c}
\toprule
& & \multicolumn{3}{c}{\textbf{(a) }\(\mathbf{R}_{\text{MBPP+}}\)} & \multicolumn{3}{c}{\textbf{(b) }\(\mathbf{R}_{\text{MATH-500}}\)} \\
\cmidrule(lr){3-5} \cmidrule(lr){6-8}
\textbf{Metric} & \textbf{Module} & {\textbf{MBPP+\indistmark}} & {\textbf{MATH-500}} & {\textbf{WikiText-2}} & {\textbf{MBPP+}} & {\textbf{MATH-500\indistmark}} & {\textbf{WikiText-2}} \\
\midrule
\multirow{2}{*}{\(\Delta_\mathrm{PPL}\)}
 & ATTN & \multicolumn{1}{S[table-format=+3.2,retain-explicit-plus,table-space-text-post={\%}]}{+30.16\%} & \multicolumn{1}{S[table-format=+3.2,retain-explicit-plus,table-space-text-post={\%}]}{+25.97\%} & \multicolumn{1}{S[table-format=+3.2,retain-explicit-plus,table-space-text-post={\%}]}{+5.82\%} & \multicolumn{1}{S[table-format=+3.2,retain-explicit-plus,table-space-text-post={\%}]}{+28.43\%} & \multicolumn{1}{S[table-format=+3.2,retain-explicit-plus,table-space-text-post={\%}]}{+27.50\%} & \multicolumn{1}{S[table-format=+3.2,retain-explicit-plus,table-space-text-post={\%}]}{+4.84\%} \\
 & MLP  & \multicolumn{1}{S[table-format=+3.2,retain-explicit-plus,table-space-text-post={\%}]}{+263.82\%} & \multicolumn{1}{S[table-format=+3.2,retain-explicit-plus,table-space-text-post={\%}]}{+170.24\%} & \multicolumn{1}{S[table-format=+3.2,retain-explicit-plus,table-space-text-post={\%}]}{+740.63\%} & \multicolumn{1}{S[table-format=+3.2,retain-explicit-plus,table-space-text-post={\%}]}{+16.86\%} & \multicolumn{1}{S[table-format=+3.2,retain-explicit-plus,table-space-text-post={\%}]}{+34.57\%} & \multicolumn{1}{S[table-format=+3.2,retain-explicit-plus,table-space-text-post={\%}]}{+58.56\%} \\
\midrule
\multirow{2}{*}{\(\bar{D}_{\mathrm{KL}}\)}
 & ATTN & \multicolumn{1}{S[table-format=1.4]}{0.2913} & \multicolumn{1}{S[table-format=1.4]}{0.2793} & \multicolumn{1}{S[table-format=1.4]}{0.2036} & \multicolumn{1}{S[table-format=1.4]}{0.2793} & \multicolumn{1}{S[table-format=1.4]}{0.2823} & \multicolumn{1}{S[table-format=1.4]}{0.2027} \\
 & MLP  & \multicolumn{1}{S[table-format=1.4]}{1.4031} & \multicolumn{1}{S[table-format=1.4]}{1.1246} & \multicolumn{1}{S[table-format=1.4]}{2.4549} & \multicolumn{1}{S[table-format=1.4]}{0.1721} & \multicolumn{1}{S[table-format=1.4]}{0.3280} & \multicolumn{1}{S[table-format=1.4]}{0.3476} \\
\bottomrule
\end{tabular}
}
\caption{
w/o-RSF domain perplexity degradation (\(\Delta_\mathrm{PPL}\), Eq.~\ref{eq:delta_ppl}) and mean per-token forward KL divergence (\(\bar{D}_{\mathrm{KL}}\), Eq.~\ref{eq:mean_kl}) on Qwen3-4B-it after suppressing the top \(K_{\mathrm{sel}}{=}5\) RACE-selected neurons per layer.
Columns report interventions under \(\mathbf{R}_{\text{MBPP+}}\) and \(\mathbf{R}_{\text{MATH-500}}\); all evaluations use direct model comparison against the unmodified model.}\label{tab:appendix_no_rsf_ppl_kl_qwen3}
\end{table}

\section{Code Domain Intervention with Reference-Set Filtering on Qwen3-4B-it}
\label{sec:appendix_code_reference_filtered_qwen3}

This appendix reports the code-domain Qwen3-4B-it results summarized by the radar plot in Figure~\ref{fig:radar_code_math_suppression}.
Neurons are selected with \(\mathbf{R}_{\text{MBPP+} \setminus \text{WikiText-2}}\), and Table~\ref{tab:appendix_code_suppression_reference_filtered_qwen3} provides the full benchmark and ISI values.

\begin{table}[ht]
\centering
\scriptsize
\setlength{\tabcolsep}{3pt}
\resizebox{\linewidth}{!}{
\begin{tabular}{c@{\hspace{3pt}}l@{\hspace{3pt}}S[table-format=2.2]@{\hspace{5pt}}S[table-format=2.2]@{\hspace{5pt}}S[table-format=2.2]@{\hspace{5pt}}S[table-format=2.2]@{\hspace{5pt}}S[table-format=2.2]@{\hspace{5pt}}S[table-format=2.2]@{\hspace{5pt}}S[table-format=1.2]}
\toprule
\textbf{Module} & \textbf{Method} & {\textbf{MBPP+\indistmark} \(\downarrow\)} & {\textbf{HumanEval+\oodmark} \(\downarrow\)} & {\textbf{MATH-500}} & {\textbf{AMC}} & {\textbf{MMLU-Redux}} & {\textbf{GPQA}} & {\textbf{ISI}\(\uparrow\)} \\
\midrule
\multirow{7}{*}{ATTN}
 & GxAct & 8.73 & 7.93 & 17.20 & 2.24 & 47.67 & 21.72 & 0.26 \\
 & AttnLRP & 0.00 & 0.00 & 2.60 & 0.75 & 6.07 & 5.05 & 0.05 \\
 & Act. Mean & 81.75 & 86.59 & 90.03 & 75.37 & 80.67 & 46.97 & 0.00 \\
 & Emp. Mean & 66.40 & 78.66 & 88.34 & 63.43 & 78.54 & 40.40 & 0.00 \\
 & Emp. SNR & 68.52 & 84.15 & 93.44 & 76.86 & 79.49 & 40.40 & 0.09 \\
 & Neg. CAM & 81.75 & 82.93 & 92.61 & 83.58 & 81.53 & 45.45 & 0.00 \\
\rowcolor[RGB]{236,244,252}\cellcolor{white} & \textbf{RACE} & 65.87 & 80.49 & 90.30 & 68.66 & 79.07 & 44.41 & 0.29 \\
\midrule
\multirow{7}{*}{MLP}
 & GxAct & 23.81 & 19.51 & 79.23 & 46.27 & 78.49 & 40.40 & 1.29 \\
 & AttnLRP & 0.00 & 3.05 & 61.37 & 35.07 & 76.39 & 31.82 & 1.07 \\
 & Act. Mean & 11.38 & 12.21 & 88.73 & 72.09 & 82.61 & 40.98 & 2.22 \\
 & Emp. Mean & 5.82 & 2.44 & 86.44 & 63.43 & 82.30 & 55.05 & 2.26 \\
 & Emp. SNR & 83.33 & 84.76 & 93.83 & 69.40 & 78.98 & 38.89 & 0.00 \\
 & Neg. CAM & 68.25 & 84.76 & 93.23 & 81.34 & 81.72 & 48.99 & 1.04 \\
\rowcolor[RGB]{236,244,252}\cellcolor{white} & \textbf{RACE} & 3.17 & 1.22 & 85.41 & 75.11 & 82.35 & 52.53 & 2.65 \\
\midrule
\multicolumn{2}{l}{Qwen3-4B-it} & 82.28 & 83.54 & 94.40 & 87.31 & 81.37 & 45.45 & {---} \\
\bottomrule
\end{tabular}
}
\caption{
Code domain with \(\mathbf{R}_{\text{MBPP+} \setminus \text{WikiText-2}}\): accuracy (\%) and ISI on Qwen3-4B-it after suppressing top-\(1\%\) ATTN or MLP target-selected neurons per layer.}\label{tab:appendix_code_suppression_reference_filtered_qwen3}
\end{table}

\section{Math Domain Intervention on Qwen3-4B-it}
\label{sec:appendix_math_suppression_qwen3}

This appendix reports additional math-domain suppression results on Qwen3-4B-it in Tables~\ref{tab:appendix_math_suppression_qwen3} and~\ref{tab:appendix_math_suppression_reference_filtered_qwen3}.

\begin{table}[ht]
\centering
\scriptsize
\setlength{\tabcolsep}{3pt}
\resizebox{\linewidth}{!}{
\begin{tabular}{c@{\hspace{3pt}}l@{\hspace{3pt}}S[table-format=2.2]@{\hspace{5pt}}S[table-format=2.2]@{\hspace{5pt}}S[table-format=2.2]@{\hspace{5pt}}S[table-format=2.2]@{\hspace{5pt}}S[table-format=2.2]@{\hspace{5pt}}S[table-format=2.2]@{\hspace{5pt}}S[table-format=1.2]}
\toprule
\textbf{Module} & \textbf{Method} & {\textbf{MATH-500\indistmark}} & {\textbf{AMC\oodmark}} & {\textbf{MBPP+}} & {\textbf{HumanEval+}} & {\textbf{MMLU-Redux}} & {\textbf{GPQA}} & {\textbf{ISI}\(\uparrow\)} \\
\midrule
\multirow{7}{*}{ATTN}
 & GxAct & 80.00 & 47.01 & 59.79 & 49.39 & 78.39 & 37.37 & 0.27 \\
 & AttnLRP & 89.60 & 73.88 & 48.15 & 47.56 & 76.98 & 43.94 & 0.00 \\
 & Act. Mean & 92.80 & 86.57 & 70.11 & 63.41 & 79.54 & 45.96 & 0.00 \\
 & Emp. Mean & 93.60 & 86.56 & 72.22 & 56.71 & 79.14 & 44.95 & 0.00 \\
 & Emp. SNR & 94.20 & 84.33 & 76.72 & 49.39 & 80.07 & 41.41 & 0.00 \\
 & Neg. CAM & 94.20 & 85.82 & 73.28 & 83.54 & 79.56 & 43.94 & 0.00 \\
\rowcolor[RGB]{236,244,252}\cellcolor{white} & \textbf{RACE} & 94.20 & 84.33 & 73.28 & 51.83 & 80.18 & 42.42 & 0.00 \\
\midrule
\multirow{7}{*}{MLP}
 & GxAct & 89.60 & 70.15 & 80.42 & 71.34 & 80.07 & 41.41 & 0.46 \\
 & AttnLRP & 90.40 & 80.60 & 83.60 & 83.54 & 81.42 & 43.43 & 1.04 \\
 & Act. Mean & 92.00 & 80.60 & 78.57 & 71.95 & 81.19 & 53.03 & 0.00 \\
 & Emp. Mean & 57.40 & 35.08 & 78.31 & 85.37 & 80.30 & 42.42 & 2.47 \\
 & Emp. SNR & 93.80 & 80.60 & 71.96 & 83.54 & 80.30 & 42.93 & 0.00 \\
 & Neg. CAM & 56.00 & 35.07 & 79.63 & 85.37 & 79.98 & 40.40 & 2.31 \\
\rowcolor[RGB]{236,244,252}\cellcolor{white} & \textbf{RACE} & 63.60 & 29.11 & 79.89 & 85.98 & 80.14 & 44.44 & 2.93 \\
\midrule
\multicolumn{2}{l}{Qwen3-4B-it} & 94.40 & 87.31 & 82.28 & 83.54 & 81.37 & 45.45 & {---} \\
\bottomrule
\end{tabular}
}
\caption{
Math domain with \(\mathbf{R}_{\text{MATH-500}}\): accuracy (\%) on Qwen3-4B-it after suppressing the top \(K_{\mathrm{sel}}{=}5\) neurons per layer.}\label{tab:appendix_math_suppression_qwen3}
\end{table}

\begin{table}[ht]
\centering
\scriptsize
\setlength{\tabcolsep}{3pt}
\resizebox{\linewidth}{!}{
\begin{tabular}{c@{\hspace{3pt}}l@{\hspace{3pt}}S[table-format=2.2]@{\hspace{5pt}}S[table-format=2.2]@{\hspace{5pt}}S[table-format=2.2]@{\hspace{5pt}}S[table-format=2.2]@{\hspace{5pt}}S[table-format=2.2]@{\hspace{5pt}}S[table-format=2.2]@{\hspace{5pt}}S[table-format=1.2]}
\toprule
\textbf{Module} & \textbf{Method} & {\textbf{MATH-500\indistmark}} & {\textbf{AMC\oodmark}} & {\textbf{MBPP+}} & {\textbf{HumanEval+}} & {\textbf{MMLU-Redux}} & {\textbf{GPQA}} & {\textbf{ISI}\(\uparrow\)} \\
\midrule
\multirow{7}{*}{ATTN}
 & GxAct & 4.00 & 2.24 & 1.06 & 0.00 & 16.98 & 7.07 & 0.05 \\
 & AttnLRP & 1.00 & 1.49 & 0.00 & 0.00 & 4.33 & 2.53 & 0.00 \\
 & Act. Mean & 73.20 & 41.60 & 78.04 & 78.66 & 75.23 & 36.87 & 1.28 \\
 & Emp. Mean & 92.20 & 73.88 & 69.31 & 83.54 & 78.33 & 40.40 & 0.02 \\
 & Emp. SNR & 91.80 & 78.36 & 82.54 & 84.76 & 79.61 & 43.94 & 1.01 \\
 & Neg. CAM & 91.20 & 82.84 & 79.89 & 79.88 & 81.26 & 42.93 & 0.00 \\
\rowcolor[RGB]{236,244,252}\cellcolor{white} & \textbf{RACE} & 90.60 & 72.39 & 71.43 & 84.76 & 78.54 & 40.91 & 0.32 \\
\midrule
\multirow{7}{*}{MLP}
 & GxAct & 30.80 & 9.70 & 69.02 & 51.39 & 77.19 & 41.41 & 1.46 \\
 & AttnLRP & 46.00 & 31.34 & 53.70 & 33.54 & 79.16 & 43.43 & 0.78 \\
 & Act. Mean & 58.40 & 47.76 & 65.61 & 48.17 & 79.79 & 40.91 & 0.75 \\
 & Emp. Mean & 27.20 & 12.69 & 61.64 & 55.49 & 77.74 & 39.39 & 1.36 \\
 & Emp. SNR & 95.00 & 86.57 & 74.07 & 82.93 & 80.70 & 39.90 & 0.00 \\
 & Neg. CAM & 95.60 & 85.08 & 80.69 & 85.98 & 81.44 & 44.44 & 0.00 \\
\rowcolor[RGB]{236,244,252}\cellcolor{white} & \textbf{RACE} & 29.00 & 11.94 & 70.20 & 67.49 & 77.79 & 40.40 & 1.76 \\

\midrule
\multicolumn{2}{l}{Qwen3-4B-it} & 94.40 & 87.31 & 82.28 & 83.54 & 81.37 & 45.45 & {---} \\
\bottomrule
\end{tabular}
}
\caption{
Math domain with \(\mathbf{R}_{\text{MATH-500} \setminus \text{WikiText-2}}\): accuracy (\%) on Qwen3-4B-it after suppressing top-\(1\%\) ATTN or MLP target-selected neurons per layer.}\label{tab:appendix_math_suppression_reference_filtered_qwen3}
\end{table}

\section{Bayesian Variance Regularization in Low-Data Regimes}
\label{sec:appendix_variance_regularization}

This appendix expands the variance-regularization argument summarized in \S\ref{subsec:discussion}.
Let \(N=|D_c|\) denote the number of input samples.
Let \(n\) denote the token-position observation count defined in \S\ref{subsec:problem_formulation}.
Under the default RACE prior \(\mu_0=0\), \(\lambda_0=1\), \(\alpha_0=1\), and \(\beta_0=1\), the NIG update in Eq.~\eqref{eq:nig_beta} becomes
\begin{equation}
    \beta_{n,j}
    =
    1+\frac{\mathrm{SS}_j}{2}+\frac{n\bar{e}_j^2}{2(1+n)}
    \ge 1.
\end{equation}
This lower bound holds even when the empirical dispersion collapses to \(\mathrm{SS}_j=0\).
Since \(\alpha_n=1+n/2\) and \(\lambda_n=1+n\), the posterior scale for mean evidence satisfies
\begin{equation}
    \sigma_{\mu,j}
    =
    \sqrt{\frac{\beta_{n,j}}{\alpha_n\lambda_n}}
    \ge
    \sqrt{\frac{1}{(1+n/2)(1+n)}} > 0.
\end{equation}
Thus CAM retains a finite conservative margin against weak or sparsity-induced evidence at finite \(n\).
This prevents near-zero empirical variance from eliminating the uncertainty penalty.
As \(n\) grows, this lower bound decays, matching the main-text discussion that the prior becomes negligible in large-\(n\) regimes.

\section{Suppression-Budget Sweep}
\label{sec:appendix_suppression_budget_sweep}

This appendix reports how the intervention effect evolves as the per-layer suppression budget varies, complementing the fixed-budget results in \S\ref{subsubsec:llm_suppression}.
We sweep the proportional per-layer MLP suppression budget \(\tau_{\mathrm{sel}}\in\{0.10\%,0.25\%,0.50\%,0.75\%,1.00\%\}\) under the Qwen3-4B-it code-RSF setting: neurons are selected with \(\mathbf{R}_{\text{MBPP+} \setminus \text{WikiText-2}}\), and the top \(\tau_{\mathrm{sel}}\) fraction of MLP neurons within every layer is suppressed.
Table~\ref{tab:appendix_budget_sweep} reports the resulting code-target accuracies.

\begin{table}[ht]
\centering
\scriptsize
\setlength{\tabcolsep}{6pt}
\begin{tabular}{lcc}
\toprule
\textbf{Budget} & \textbf{HumanEval+\oodmark} & \textbf{MBPP+\indistmark} \\
\midrule
None        & 83.54 & 82.28 \\
Top \(0.10\%\) & 82.93 & 75.93 \\
Top \(0.25\%\) & 75.61 & 64.29 \\
Top \(0.50\%\) & 40.24 & 59.26 \\
Top \(0.75\%\) & 23.17 & 33.86 \\
Top \(1.00\%\) & 1.22  & 3.17  \\
\bottomrule
\end{tabular}
\caption{
Suppression-budget sweep on Qwen3-4B-it under \(\mathbf{R}_{\text{MBPP+} \setminus \text{WikiText-2}}\): accuracy (\%) on the two code benchmarks as the per-layer MLP suppression budget varies.
``None'' denotes the unmodified model.}
\label{tab:appendix_budget_sweep}
\end{table}

The effect strengthens progressively with the budget: degradation is mild at \(0.10\%\), becomes substantial from \(0.25\%\) onward, and approaches the floor only at the largest budget.
The graded response across budgets places the near-floor outcome at the top-\(1\%\) budget at the end of a continuous trend.

\section{Ranking Stability under Scoring-Set Subsampling}
\label{sec:appendix_ranking_stability}

This appendix measures how the selected neuron sets converge as the size of the target-domain scoring set grows.
Using seed \(42\), we construct nested MBPP+ subsets so that every larger run retains all examples from the preceding smaller run.
For each subset size \(N\) and selection fraction \(\tau_{\mathrm{sel}}\in\{1\%,5\%,10\%\}\), we score neurons on the subset and compare the resulting selected sets with those obtained from the full MBPP+ scoring set (\(N{=}378\)), separately for the MLP and ATTN modules.
We report the macro-averaged Jaccard overlap, where the macro average is taken over layers within each module.

\begin{table}[ht]
\centering
\scriptsize
\setlength{\tabcolsep}{5pt}
\resizebox{\columnwidth}{!}{
\begin{tabular}{rccc@{\hspace{14pt}}ccc}
\toprule
& \multicolumn{3}{c}{\textbf{MLP Jaccard}} & \multicolumn{3}{c}{\textbf{ATTN Jaccard}} \\
\cmidrule(lr){2-4} \cmidrule(lr){5-7}
\textbf{\(N\)} & {Top-\(1\%\)} & {Top-\(5\%\)} & {Top-\(10\%\)} & {Top-\(1\%\)} & {Top-\(5\%\)} & {Top-\(10\%\)} \\
\midrule
10  & 0.58 & 0.64 & 0.67 & 0.71 & 0.71 & 0.71 \\
50  & 0.71 & 0.76 & 0.79 & 0.86 & 0.83 & 0.83 \\
100 & 0.82 & 0.85 & 0.88 & 0.93 & 0.91 & 0.90 \\
200 & 0.89 & 0.91 & 0.93 & 0.95 & 0.94 & 0.94 \\
300 & 0.92 & 0.94 & 0.94 & 0.95 & 0.95 & 0.95 \\
378 & 1.00 & 1.00 & 1.00 & 1.00 & 1.00 & 1.00 \\
\bottomrule
\end{tabular}
}
\caption{
Ranking stability under scoring-set subsampling on Qwen3-4B-it.
Each entry is the macro-averaged Jaccard overlap between neuron sets selected at the same fraction \(\tau_{\mathrm{sel}}\) from a nested MBPP+ subset of size \(N\) and from the full scoring set (\(N{=}378\)).}
\label{tab:appendix_ranking_stability}
\end{table}

Table~\ref{tab:appendix_ranking_stability} shows that the overlap increases monotonically with scoring-set size in both modules.
With \(N{=}100\) examples, the top-\(1\%\) overlap already reaches \(0.82\) for MLP and \(0.93\) for ATTN; with \(N{\ge}200\), all reported overlaps exceed \(0.89\) for MLP and \(0.94\) for ATTN.
The selected neuron sets approach the full-data ranking as the scoring set grows, providing direct evidence about the amount of target-domain data needed for stable selection.

\section{Robustness to the Reference Corpus}
\label{sec:appendix_reference_corpus_robustness}

This appendix tests whether the intervention pattern depends on the specific choice of the general-text reference corpus used by RSF.
We replace WikiText-2 with the official \texttt{sample-10BT} subset of FineWeb~\citep{penedo2024fineweb}, which is randomly sampled from the full FineWeb corpus and therefore preserves broad source diversity.
Using seed \(42\), we discard texts shorter than \(80\) characters, truncate documents to at most \(2{,}000\) characters, and materialize exactly \(10{,}000\) examples as the alternative reference corpus.
We then repeat the Qwen3-4B-it code-domain experiment with MBPP+ as the scoring set, suppressing the top-\(1\%\) MLP neurons within every layer under \(\mathbf{R}_{\text{MBPP+} \setminus \text{FineWeb}}\).
Table~\ref{tab:appendix_reference_corpus_robustness} compares the resulting benchmark accuracies with the WikiText-2 reference-set results.

\begin{table}[ht]
\centering
\scriptsize
\setlength{\tabcolsep}{5pt}
\resizebox{\columnwidth}{!}{
\begin{tabular}{lcc}
\toprule
\textbf{Benchmark} & \textbf{WikiText-2 reference} & \textbf{FineWeb \texttt{sample-10BT} reference} \\
\midrule
MBPP+\indistmark   & 3.17  & 3.70  \\
HumanEval+\oodmark & 1.22  & 2.44  \\
MATH-500           & 85.41 & 84.40 \\
AMC                & 75.11 & 73.14 \\
MMLU-Redux         & 82.35 & 82.68 \\
GPQA               & 52.53 & 55.56 \\
\bottomrule
\end{tabular}
}
\caption{
Robustness to the reference corpus on Qwen3-4B-it: accuracy (\%) after per-layer top-\(1\%\) MLP suppression under \(\mathbf{R}_{\text{MBPP+} \setminus D_{\mathrm{ref}}}\), with \(D_{\mathrm{ref}}\) instantiated by WikiText-2 or by FineWeb \texttt{sample-10BT}.}
\label{tab:appendix_reference_corpus_robustness}
\end{table}

Changing the reference corpus alters each benchmark by at most \(3.03\) percentage points, with a mean absolute difference of \(1.35\) points.
The similar target-versus-control profiles obtained with both corpora show that the downstream intervention pattern is robust to the choice of reference corpus.

\section{End-to-End Runtime and Memory Footprint}
\label{sec:appendix_runtime_memory}

This appendix reports a wall-clock and memory measurement of the full RACE pipeline on a single NVIDIA RTX 4090, complementing the profiler-based FLOP analysis in Appendix~\ref{sec:appendix_profiler_scoring_cost}.
For Qwen3-4B-it on MBPP+ (scoring set \(N{=}378\) prompts, full forward evidence logging, RDA, and CAM), the end-to-end runtime is \(587.28\)\,s.
The peak reserved GPU memory is, to within measurement noise, identical to running the same forward pass with the HuggingFace \texttt{transformers} model alone: RACE adds no measurable extra allocation beyond the forward-pass weights and activations.

The only persistent state RACE maintains is one scalar Bayesian sufficient-statistic record per native neuron, which can be kept on CPU and streamed to GPU only for the reduction.
Even when kept on GPU, its footprint is negligible: Qwen3-4B-it has \(497{,}664\) native neurons, so one fp16 RACE variable per neuron costs \(497{,}664 \times 2\,\mathrm{B} \approx 0.95\)\,MB, versus \(\approx\)\(8\)\,GB for the model weights alone---roughly \(0.01\%\) of the weight footprint and orders of magnitude below forward-pass activation memory.
The RACE state grows linearly in the number of neurons and hence in model size.
Its streaming memory remains constant as corpus length increases.
Runtime grows linearly with the number of forward passes.

\end{document}